\documentclass[letterpaper]{article} 
\usepackage[draft]{aaai2026}  
\usepackage{times}  
\usepackage{helvet}  
\usepackage{courier}  
\usepackage[hyphens]{url}  
\usepackage{graphicx} 
\usepackage{natbib}  
\usepackage{caption} 
\usepackage{algorithm}
\usepackage{algorithmic}

\usepackage[T1]{fontenc}
\usepackage{upquote}

\usepackage{newfloat}
\usepackage{listings}
\DeclareCaptionStyle{ruled}{labelfont=normalfont,labelsep=colon,strut=off} 
\floatstyle{ruled}
\newfloat{listing}{tb}{lst}{}
\floatname{listing}{Listing}
\usepackage[most]{tcolorbox}

\definecolor{bluecite}{HTML}{0875b7}

\newtcolorbox{mybox}[1]{%
  colback=bluecite!5!white,
  colframe=bluecite!75!black,
  fonttitle=\bfseries,
  title={#1}
}

\usepackage{latexsym}
\usepackage{amssymb}
\usepackage{mathtools}
\usepackage{amsmath}
\usepackage{amsthm}
\usepackage{microtype}
\usepackage{booktabs, array}
\usepackage{enumitem}
\usepackage{graphicx} 
\usepackage{color} 
\usepackage{hyperref}

\usepackage{subcaption}

\usepackage[table]{xcolor}

\usepackage{rotating}
\usepackage{nicefrac}
\usepackage{siunitx}

\usepackage{amsfonts}
\usepackage{multirow}

\usepackage{pifont}

\newcommand{\ffn}[2]{%
  \refstepcounter{footnote}%
  \label{#1}%
  \textsuperscript{\thefootnote} #2\par
}

\newcommand{\rotlabel}[1]{\rotatebox{50}{#1}}

\newcommand{\BibTeX}{B\kern-.05em{\sc i\kern-.025em b}\kern-.08em\TeX}

\definecolor{newdarkgreen}{RGB}{0,90,96}
\definecolor{newlightgreen}{RGB}{47,190,173}
\definecolor{newyellow}{RGB}{210,166,131}
\definecolor{newred}{RGB}{193,67,60}
\definecolor{newpink}{RGB}{255,137,133}

\newcommand{\cm}{\textcolor{newdarkgreen}{\ding{51}}} 
\newcommand{\xm}{\textcolor{newred}{\ding{55}}}

\definecolor{bestl}{HTML}{2FBEAD}

\usepackage{array}
\newcolumntype{L}[1]{>{\raggedright\arraybackslash}p{#1}} 
\newcolumntype{C}[1]{>{\centering\arraybackslash}p{#1}}   

\definecolor{best}{HTML}{E8D5C4}
\definecolor{bestl}{HTML}{2FBEAD}

\newcommand{\compactbest}[1]{\setlength{\fboxsep}{1pt}\colorbox{best!100}{#1}}

\title{CAMAR: Continuous Actions Multi-Agent Routing}
\author {
    Artem Pshenitsyn\textsuperscript{\rm 1,2},
    Aleksandr Panov\textsuperscript{\rm 1,2},
    Alexey Skrynnik\textsuperscript{\rm 1,2}
}
\affiliations {
    \textsuperscript{\rm 1}CogAI Lab, Moscow, Russia\\
    \textsuperscript{\rm 2}MIRAI, Moscow, Russia\\
    pshenitsyn@cogailab.com,
    skrynnikalexey@gmail.com
}

\usepackage{bibentry}
\begin{document}

\maketitle

\begin{abstract}

Multi-agent reinforcement learning (MARL) is a powerful paradigm for solving cooperative and competitive decision-making problems. While many MARL benchmarks have been proposed, few combine continuous state and action spaces with challenging coordination and planning tasks. We introduce CAMAR, a new MARL benchmark designed explicitly for multi-agent pathfinding in environments with continuous actions. CAMAR supports cooperative and competitive interactions between agents and runs efficiently at up to 100,000 environment steps per second. We also propose a three-tier evaluation protocol to better track algorithmic progress and enable deeper analysis of performance. In addition, CAMAR allows the integration of classical planning methods such as RRT and RRT* into MARL pipelines. We use them as standalone baselines and combine RRT* with popular MARL algorithms to create hybrid approaches. We provide a suite of test scenarios and benchmarking tools to ensure reproducibility and fair comparison. Experiments show that CAMAR presents a challenging and realistic testbed for the MARL community.
\end{abstract}

\begin{links}
\small
\link{Code}{https://github.com/AIRI-Institute/CAMAR.git}
\end{links}


\section{Introduction}

Multi-agent reinforcement learning (MARL) has shown strong results in cooperative and competitive settings, and many recent studies explore how MARL can solve tasks that need coordination in complex environments~\cite{de2020independent, bettini2023heterogeneous, yu2022surprising, damani2021primal, skrynnik2024learn, andreychuk2025mapf}. One important group of such tasks is multi-agent pathfinding (MAPF), where several agents must reach their goals without collisions. Classic MAPF is usually studied on discrete grids, but real robots move in continuous space, follow dynamics, and must plan smooth paths. This continuous version of MAPF is important for many domains such as warehouse logistics, drone swarm coordination, and other systems where many robots must move safely and efficiently.

\begin{figure}[t]
    \centering
    \includegraphics[width=1.0\linewidth]{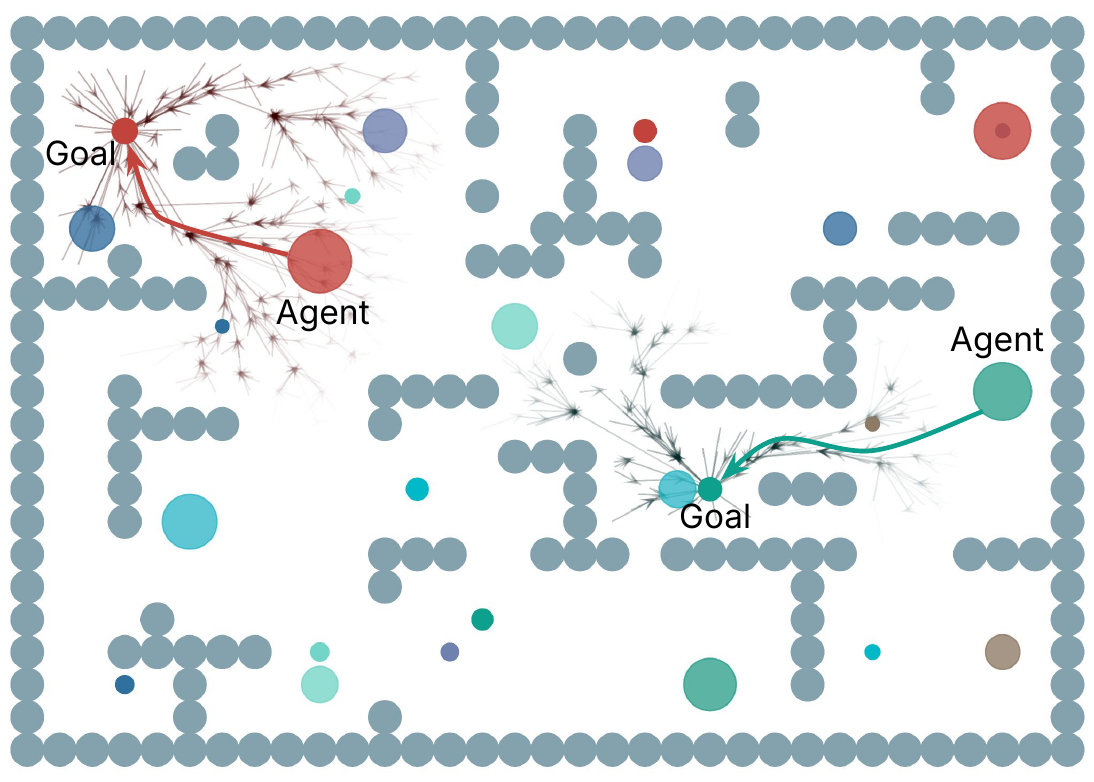}

    \caption{An example scenario from the proposed CAMAR benchmark. Agents are represented as filled circles. Each agent aims to reach its goal while avoiding collisions. The small arrows for the red and green agents indicate segments of paths generated by RRT*, providing guidance for the RL algorithms.}
    \label{fig:env_overview}

\end{figure}

Learnable methods have recently become effective for MAPF and cooperative navigation~\cite{angulo2022policy, skrynnik2024learn, andreychuk2025mapf, damani2021primal}. However, many existing environments still use grid worlds or discrete actions that do not match real robot behavior. In practice, robots must plan and coordinate while avoiding both static and dynamic obstacles, a long-standing problem in robotics~\cite{chen2024survey, wang2020mobile, angulo2022policy, lehoux2024multi}. Many MARL environments simplify this problem too much by using low-dimensional maps or unrealistic movement models.

High-fidelity simulators like Gazebo\footnote{\url{https://gazebosim.org/home}}, Isaac Sim\footnote{\url{https://developer.nvidia.com/isaac/sim}}, AirSim~\cite{shah2017airsim}, and Flightmare~\cite{song2021flightmare} offer realistic physics, but they focus on robot control and perception, not on large-scale MARL. These tools often run slower and cannot simulate hundreds of agents at once. To study large-scale continuous navigation, we need environments that combine high speed with continuous dynamics, run efficiently on GPU, and allow testing both learned and planning-based methods in a single benchmark.

We identify three main gaps in current MARL environments. First, many use discrete action spaces that cannot represent smooth motion~\cite{papoudakis2020benchmarking, skrynnik2022pogema, skrynnik2025pogema}. Second, while some environments do support continuous states and actions, they do not scale to large numbers of agents and obstacles~\cite{bettini2022vmas}. Third, other environments can scale but offer simple tasks that do not require strong coordination or realistic navigation skills~\cite{ellis2023smacv2, lowe2017multi, rutherford2023jaxmarl}.

To bridge the gap between multi-robot systems and MARL research, we introduce the CAMAR (Continuous Actions Multi-Agent Routing) Benchmark. Specifically, we make the following contributions:

\begin{itemize}
    \item We introduce CAMAR, an extremely fast environment with GPU acceleration support (using JAX), achieving speeds exceeding 100,000 steps per second. It is designed for multi-agent navigation and collision avoidance tasks in continuous state and action spaces.
    \item We propose an evaluation protocol that includes both training and holdout task instances, as well as a suite of metrics and performance indicators to assess agents’ generalization capabilities.
    \item We provide strong baselines for benchmarking, including state-of-the-art MARL algorithms and classical path planning methods commonly used in robotics, and conduct an extensive experimental study to evaluate their performance across diverse scenarios.
\end{itemize}

\section{Related Work}

To compare existing MARL environments, we evaluate them in Table~\ref{tab:marl-environments} across key features such as continuous control, GPU support, scalability, and usability. An extended version of the Table~\ref{tab:marl-environments} with descriptions of comparison features and detailed analysis of each environment are provided in the Appendix H.

Prominent benchmarks include SMAC~\cite{samvelyan2019starcraft, ellis2023smacv2, rutherford2023jaxmarl}, Jumanji~\cite{bonnet2023jumanji}, POGEMA~\cite{skrynnik2022pogema, skrynnik2025pogema}, MPE~\cite{mordatch2017emergence, lowe2017multi}, and VMAS~\cite{bettini2022vmas}. SMAC enables testing of strategic behavior but uses discrete actions and scales poorly. Jumanji supports GPUs and procedural generation but is not focused on navigation. POGEMA handles large-agent navigation with procedural maps but lacks continuous control. MPE, while foundational, does not scale to many agents; VMAS adds physical realism but still struggles with performance and scalability.

Many environments lack evaluation protocols and suffer from slow training due to CPU-GPU bottlenecks. Simulators like Gazebo~\cite{koenig2004design}, Webots~\cite{michel2004cyberbotics}, and ARGoS~\cite{pinciroli2012argos} offer realistic continuous dynamics but are not optimized for efficient MARL training. These gaps highlight the need for a new benchmark supporting scalable, high-performance multi-agent learning.

\begin{table*}[htb!]
\centering 
\small
\rowcolors{2}{gray!15}{white}
\begin{tabular}{p{3.8cm}p{0.6cm}p{0.3cm}p{0.3cm}p{0.3cm}p{0.3cm}p{0.3cm}p{0.3cm}p{0.88cm}p{0.3cm}p{0.3cm}p{0.88cm}p{0.3cm}p{0.3cm}p{0.3cm}}
Environment / Simulator & \rotlabel{Repository} & \rotlabel{Cont. Observations} & \rotlabel{Cont. Actions} & \rotlabel{GPU Support} & \rotlabel{Scalability \textgreater{}500 Agents} & \rotlabel{Partially observable} & \rotlabel{Heterogeneous agents} & \rotlabel{Performance \textgreater{}10K SPS}  & \rotlabel{Python based} & \rotlabel{Procedural generation} & \rotlabel{Requires generalization} & \rotlabel{Evaluation protocols} & \rotlabel{Tests \& CI} & \rotlabel{PyPI Listed} \\
\midrule



RWare (Jumanji)~\cite{bonnet2023jumanji} & \href{https://github.com/instadeepai/jumanji.git}{link} & \xm & \xm & \cm & \xm & \cm & \xm & \xm & \cm & \xm & \cm & \xm & \cm & \cm \\



SMAC~\cite{samvelyan2019starcraft} & \href{https://github.com/oxwhirl/smac}{link} & \cm & \xm & \xm & \xm & \cm & \cm & \xm & \xm & \xm & \xm & \xm & \xm & \xm \\

SMACv2~\cite{ellis2023smacv2} & \href{https://github.com/oxwhirl/smacv2.git}{link} & \cm & \xm & \xm & \xm & \cm & \cm & \xm & \xm & \xm & \cm & \xm & \xm & \xm \\

SMAX (JaxMARL)~\cite{rutherford2023jaxmarl} & \href{https://github.com/FLAIROx/JaxMARL.git}{link} & \cm & \cm & \cm & \xm & \cm & \cm & \cm & \cm & \xm & \cm & \xm & \cm & \cm \\

MPE ~\cite{mordatch2017emergence, lowe2017multi} & \href{https://github.com/Farama-Foundation/MPE2.git}{link} & \cm & \cm & \xm & \cm & \cm & \cm & \xm\ /\ \cm$^{~\ref{ff:sps}}$ & \cm & \xm & \cm & \xm & \cm & \cm \\

MPE (JaxMARL)~\cite{rutherford2023jaxmarl} & \href{https://github.com/FLAIROx/JaxMARL.git}{link} & \cm & \cm & \cm & \cm & \cm & \cm & \xm\ /\ \cm$^{~\ref{ff:sps}}$ & \cm & \xm & \cm & \xm & \cm & \cm \\



POGEMA~\cite{skrynnik2022pogema} & \href{https://github.com/CognitiveAISystems/pogema.git}{link} & \xm & \xm & \xm & \cm & \cm & \xm & \cm & \cm & \cm & \cm & \cm & \cm & \cm \\

VMAS$^{~\ref{ff:vmas}}$~\cite{bettini2022vmas} & \href{https://github.com/proroklab/VectorizedMultiAgentSimulator.git}{link} & \cm & \cm & \cm & \xm & \cm & \cm & \xm\ /\ \cm$^{~\ref{ff:vmas}}$ & \cm & \xm & \xm\ /\ \cm$^{~\ref{ff:vmas}}$ & \xm & \cm & \cm \\


\midrule

Gazebo ~\cite{koenig2004design} & \href{https://gazebosim.org/home}{link} & \cm & \cm & \cm & \xm & \cm & \cm & \xm & \xm & \xm & \xm & \xm & \cm & \xm \\

Webots ~\cite{michel2004cyberbotics} & \href{https://cyberbotics.com/}{link} & \cm & \cm & \cm & \xm & \cm & \cm & \xm & \xm & \xm & \xm & \xm & \cm & \xm \\

ARGoS ~\cite{pinciroli2012argos} & \href{https://www.argos-sim.info/}{link} & \cm & \cm & \xm & \cm & \cm & \cm & \xm & \xm & \xm & \xm & \xm & \cm & \xm \\

\midrule

CAMAR (Ours) & \href{https://github.com/AIRI-Institute/CAMAR.git}{link} & \cm & \cm & \cm & \cm & \cm & \cm & \cm & \cm & \cm & \cm & \cm & \cm & \cm \\

\bottomrule
\end{tabular}
\caption{Comparison of multi-agent reinforcement learning (MARL) environments and simulators. Each row corresponds to a specific environment or a particular implementation of it. The columns indicate key properties, including support for continuous observations and actions, GPU acceleration, scalability beyond 500 agents, partial observability, heterogeneous agents, and whether the simulator can exceed 10K simulation steps per second (SPS). Additional columns specify if the environment is implemented fully in Python, supports procedural generation, requires generalization across different maps or tasks, includes evaluation protocols, and provides built-in tests or continuous integration (CI). The “Repository” column contains links to the official source code for each environment. CAMAR, listed at the bottom, is our proposed environment.}
\label{tab:marl-environments}
\end{table*}

\section{CAMAR Environment}

CAMAR is designed for continuous-space planning tasks in multi-agent environments. In this environment, multiple agents move toward their goals while avoiding both static obstacles and other moving agents. The simulation happens in a fully continuous two-dimensional space, without any predefined grids. Agents interact by applying forces, which control their movement through a simple and computationally efficient dynamic model. This approach makes the environment more realistic and easier to scale for many agents.

\paragraph{Dynamic Model \& Action Space}
A key part of CAMAR is its collision model. Similar to MPE~\cite{mordatch2017emergence, lowe2017multi} and VMAS~\cite{bettini2022vmas}, CAMAR uses a force-based system. Agents receive repulsive forces from nearby agents and obstacles. The collision force applied to agent $i$ from object $j$ is calculated using a smooth contact model, as shown in the equation below.

\begin{eqnarray*}\label{eq:mpe_dynamic}
\begin{cases} 
\vec{f}_{ij}^{\text{collision}}(t) = f_0 \frac{\Delta \vec{x}_{ij}(t)}{\| \Delta \vec{x}_{ij}(t) \|} k \log \left(1 + e^{\frac{-\left( \| \Delta \vec{x}_{ij}(t) \| - d_{\min} \right)}{k}} \right), \\\text{if } \| \Delta \vec{x}_{ij}(t) \| < d_{\min}; \\ \\
\vec{f}_{ij}^{\text{collision}}(t) = 0, \\\text{otherwise}.
\end{cases}
\end{eqnarray*}

Here, contact force $f_0$ regulates the magnitude of the repulsive force, penetration softness $k$ controls the smoothness of the contact dynamics, $\Delta \vec{x}_{ij}(t)$ is a displacement vector between agent $i$ and an object $j$ at time $t$, $d_{\min}$ defines the minimum allowable distance before collision is activated.

When two objects overlap, the force grows smoothly without sudden changes. When there is no overlap, the collision force is zero. This smooth behavior helps keep agent movement more realistic and stable. The total collision force acting on agent $i$ is calculated by summing forces from all nearby objects: $\vec{f}i^c(t) = \sum{j} \vec{f}_{ij}^{\text{collision}}(t)$.

The full environment state is updated using the collision force and agents' actions. CAMAR supports multiple types of dynamic models. Currently, we provide two built-in models: \texttt{HolonomicDynamic} and \texttt{DiffDriveDynamic}.

\subparagraph{HolonomicDynamic}

This model is simple and similar to the one used in MPE~\cite{mordatch2017emergence, lowe2017multi}. Each agent has a position and velocity. The agent moves by applying a 2D force. The next state is calculated using the semi-implicit Euler method, as described in the equation below.

\begin{eqnarray*}\label{eq:holonomic}
\begin{cases} 
\vec{v}_i(t+dt) = (1 - \texttt{damping}) \vec{v}_i(t) + \frac{\vec{f}_i^a(t) + \vec{f}_i^c(t)}{\texttt{m}} dt \\
\vec{v}_i(t+dt) := 
\begin{cases} \vec{v}_i(t+dt), \text{if } \| \vec{v}_i(t+dt) \| < \texttt{max\_speed} \\
\frac{\vec{v}_i(t+dt)}{\| \vec{v}_i(t+dt) \|} \cdot \texttt{max\_speed}, \text{otherwise.} 
\end{cases} \\
\vec{\text{pos}}_i(t+dt) = \vec{\text{pos}}_i(t) + \vec{v}_i(t+dt) dt
\end{cases}
\end{eqnarray*}

Here, $\vec{f}_i^a(t)$ is the 2D action force of agent $i$, \texttt{damping} - scalar in the range $[0, 1)$ that controls velocity decay over time, \texttt{m} is the agent mass for applying forces, \texttt{max\_speed} regulates maximum speed of an agent preserving direction but limiting speed, $dt$ is the time step duration between updates.

\subparagraph{DiffDriveDynamic}

Differential-drive robot model is another built-in dynamic in CAMAR. Each agent has a 2D position $\vec{\text{pos}}_i(t)$ and a heading angle $\theta_i(t)$. The agent chooses a 2D action: one value for linear speed $u^a_i(t)$ and one for angular speed $\omega_i^a(t)$. The motion is updated based on this action using equation below.

\begin{figure}[b!]
\refstepcounter{figure}
\footnotesize
\footnoterule
\ffn{ff:sps}{SPS decreases gracefully with many agents and obstacles.}
\ffn{ff:vmas}{VMAS is a framework consisting of many different scenarios, while some scenarios run efficiently with a speed exceeding 10K SPS, other, complex ones don't; the same applies for generalization.}
\end{figure}

\begin{eqnarray*}\label{eq:diffdrive}
\begin{cases} 
u_i(t) = \text{clip}(u_i^a(t), -\texttt{max\_u}, \texttt{max\_u}) \\
\omega_i(t) = \text{clip}(\omega_i^a(t), -\texttt{max\_w}, \texttt{max\_w}) \\
\vec{v}_i(t) = [u_i(t) \cos(\theta_i(t)); u_i(t) \sin(\theta_i(t)] + \frac{\vec{f}^c_i(t)}{m}dt\\
\vec{\text{pos}}_i(t + dt) = \vec{\text{pos}}_i(t) + \vec{v}_i(t) dt \\
\theta_i(t+dt) = \theta_i(t) + \omega_i(t)dt
\end{cases}
\end{eqnarray*}
Here, \texttt{max\_u} and \texttt{max\_w} are constraints on agent velocities.

\paragraph{Observations}

Each agent in CAMAR receives a local observation centered around itself. The size of the observation window can be set by the user. This observation system is inspired by LIDAR sensors but avoids using ray tracing. Instead, CAMAR provides a simple and efficient vector-based observation.

\begin{figure}[t]
    \centering
    \includegraphics[clip, width=0.85\linewidth]{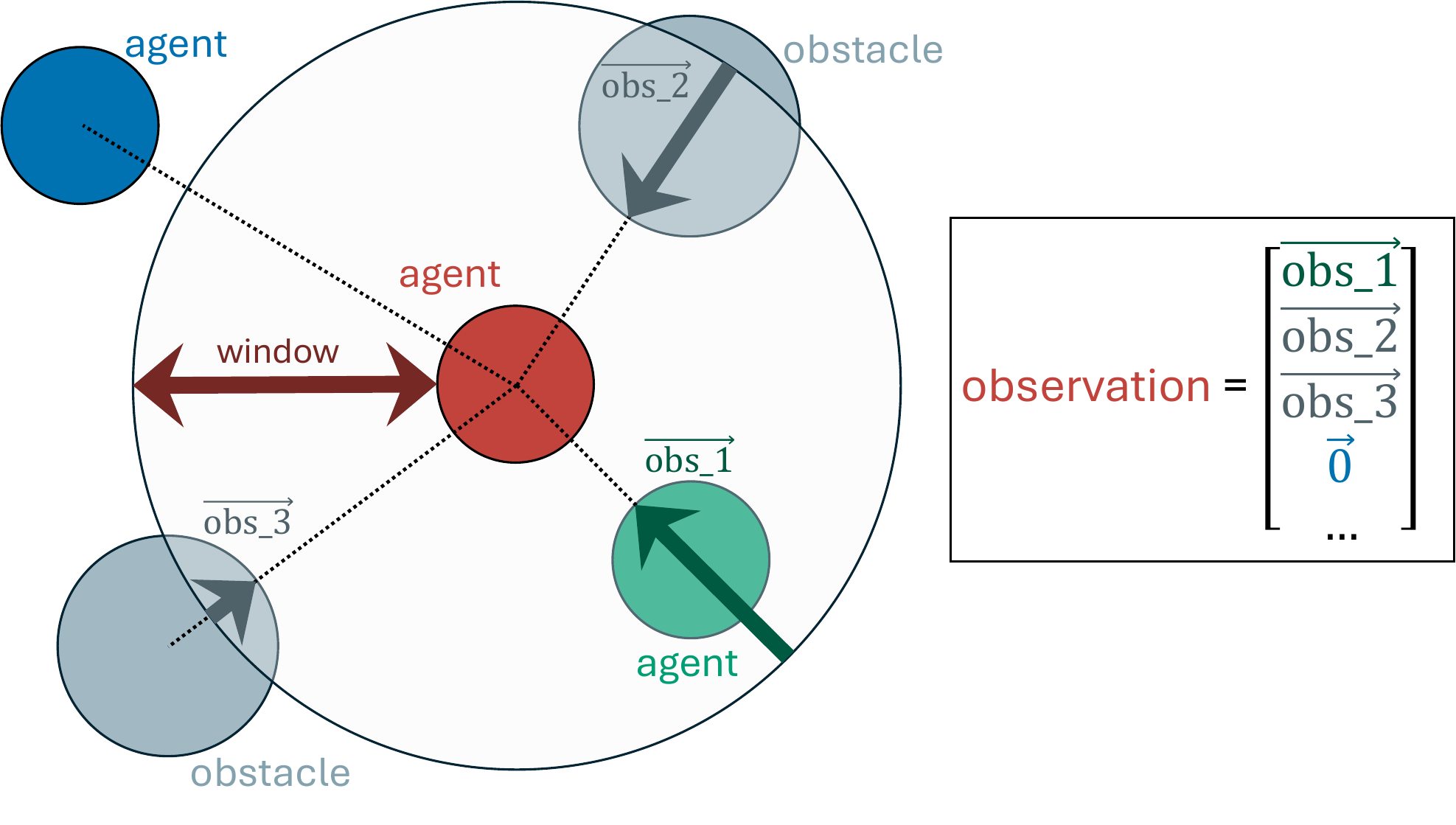}
    \caption{LIDAR-inspired vector observations in CAMAR. Each agent detects nearby objects using penetration vectors, and receives a normalized goal direction.}
    \label{fig:observation}
\end{figure}

Each agent observes nearby objects using a penetration-based vector representation, which ensures smooth and continuous observations. For every object in the environment (either an agent or a static landmark), the observation is computed as a normalized vector pointing from the agent to the object. If the object is far away, outside the agent's sensing window, the observation becomes a zero vector. This method avoids discontinuities and helps agents better generalize across different object sizes.

\begin{eqnarray*}\label{eq:observation}
\begin{cases}
\Delta \vec{o}_j = \vec{o}_j - \vec{\text{pos}} \\
\vec{\text{obs\_j}} = 
\begin{cases}
\Delta \vec{o}_j \cdot \left( 1 - \frac{\texttt{window} + R_j}{\| \vec{o}_j \|} \right), \\ \text{if } \| \Delta \vec{o}_j \| - R - R_j < \texttt{window} \\
\vec{0}, \\ \text{otherwise}.
\end{cases}\\
\vec{\text{obs\_j}} := \frac{\vec{\text{obs\_j}}}{\texttt{window}}
\end{cases}
\end{eqnarray*}

This observation is computed for each agent in a fully vectorized manner. Afterward, only the top \texttt{max\_obs} closest objects are kept to form the final observation vector.
Here, $\vec{o}_j$ is the 2D position of object $j$, $R$ is the agent radius, $R_j$ the radius of object $j$, \texttt{window} is a parameter that sets how far the agent can sense objects nearby.

In addition to obstacle information, each agent also gets an ego-centric vector pointing to its goal. This vector is clipped, normalized and concatenated to the final observation. This structure helps agents understand both their surroundings and the direction they need to move.

\paragraph{Circle-Based Discretization}
One simple way to simulate a world is to use geometry rules to check if objects overlap. Ray tracing is a common method for detecting collisions based on the shapes of objects. However, ray tracing is hard to implement in a way that is fast on a GPU. Simple versions of ray tracing are slow and better suited for CPU simulations.

Another method is to discretize the world and only check collisions with nearby objects. Based on the dynamic model described above, it is enough to calculate the distance between objects.

In CAMAR, every object is represented as a circle. This choice has several advantages. Checking the distance between two circles is simple and fast. It does not require special cases like ray tracing does. It also avoids the complexity of handling different shapes like rectangles or polygons. Because of this, the simulation can easily run on GPUs with many agents at the same time. This design allows CAMAR to simulate large-scale multi-agent tasks efficiently and with high performance.

\paragraph{Map Generators}

\begin{figure*}[ht!]
    \centering
    \begin{subfigure}{80pt}
        \centering
        \includegraphics[width=80pt]{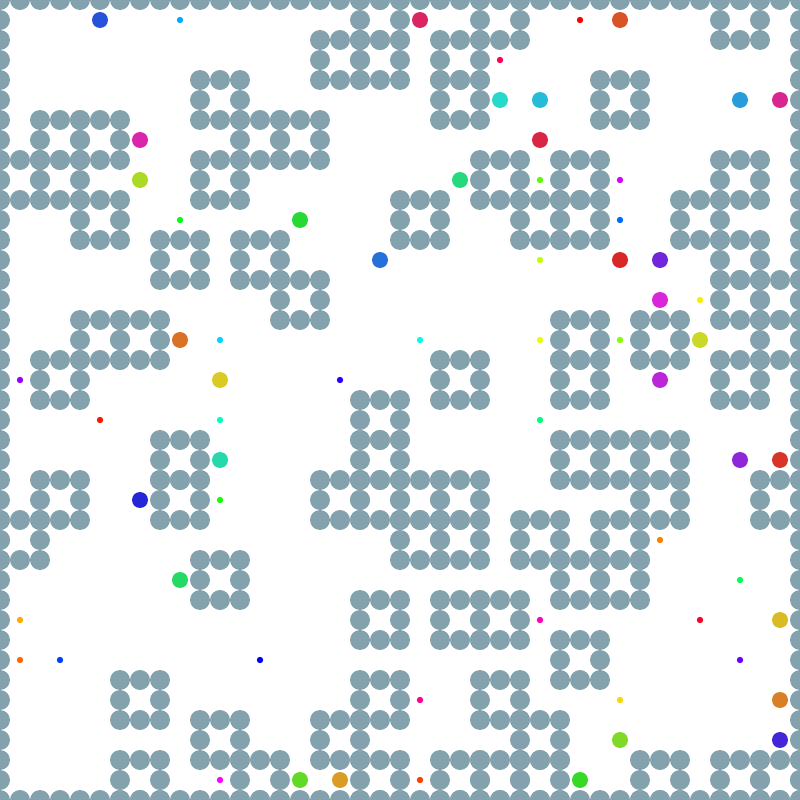}
        \caption{\texttt{random\_grid}}
        \label{subfig:maps_random_grid}
    \end{subfigure}
    \begin{subfigure}{80pt}
        \centering
        \includegraphics[width=80pt]{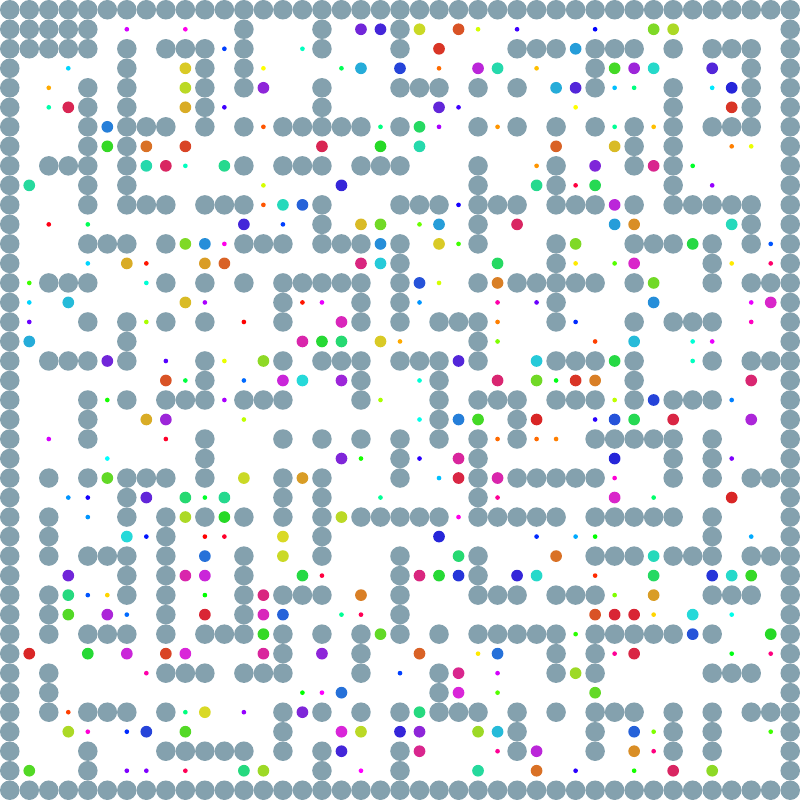}
        \caption{\texttt{labmaze\_grid}}
        \label{subfig:maps_labmaze_grid}
    \end{subfigure}
    \begin{subfigure}{80pt}
        \centering
        \includegraphics[width=80pt]{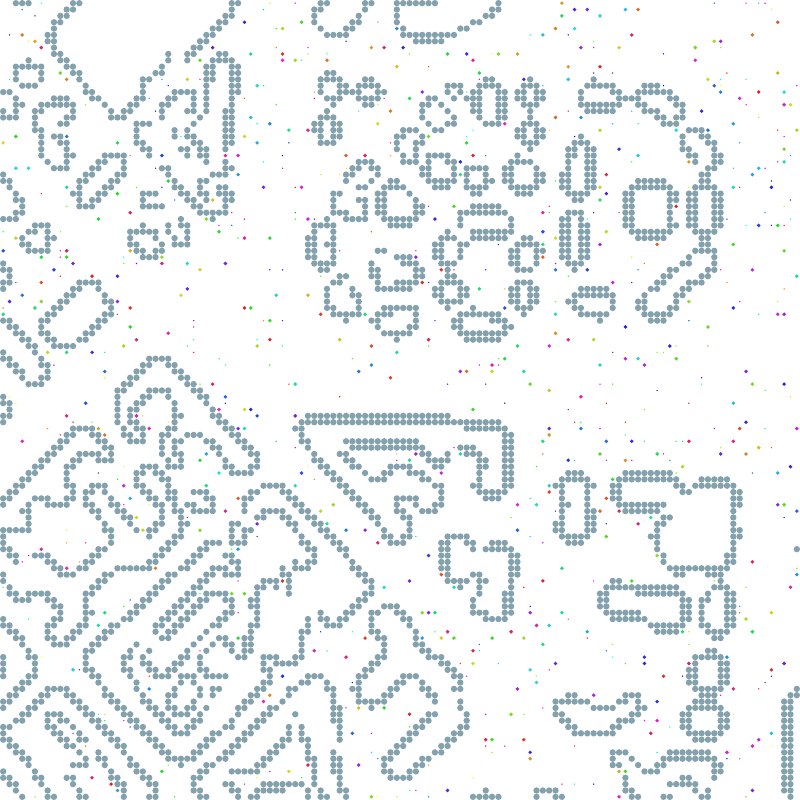}
        \caption{\texttt{movingai}}
        \label{subfig:maps_movingai}
    \end{subfigure}
    \begin{subfigure}{80pt}
        \centering
        \includegraphics[width=80pt]{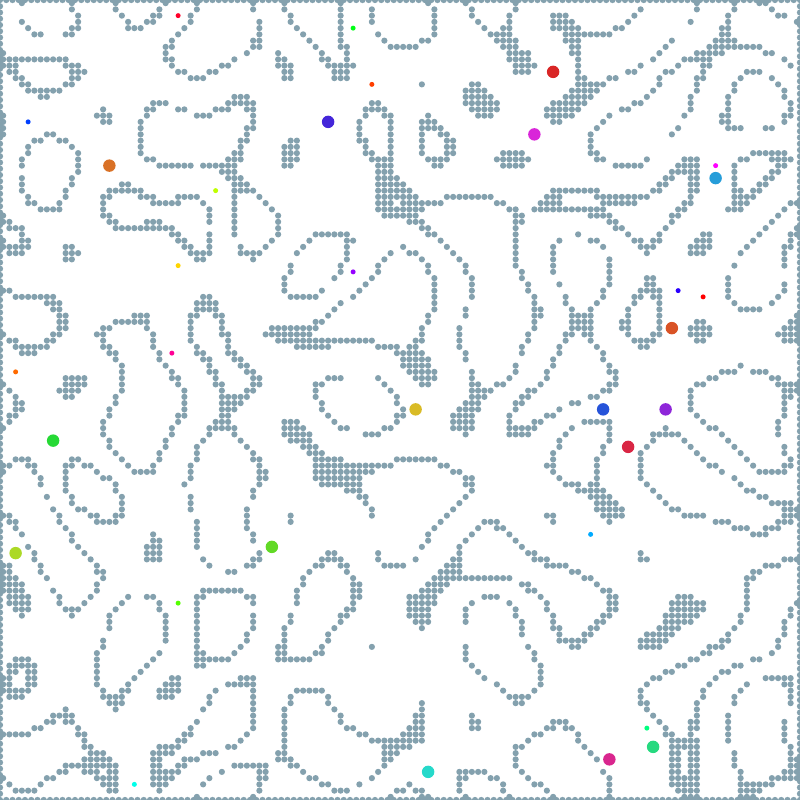}
        \caption{\texttt{caves\_cont}}
        \label{subfig:maps_caves_cont}
    \end{subfigure}
    \begin{subfigure}{80pt}
        \centering
        \includegraphics[width=80pt]{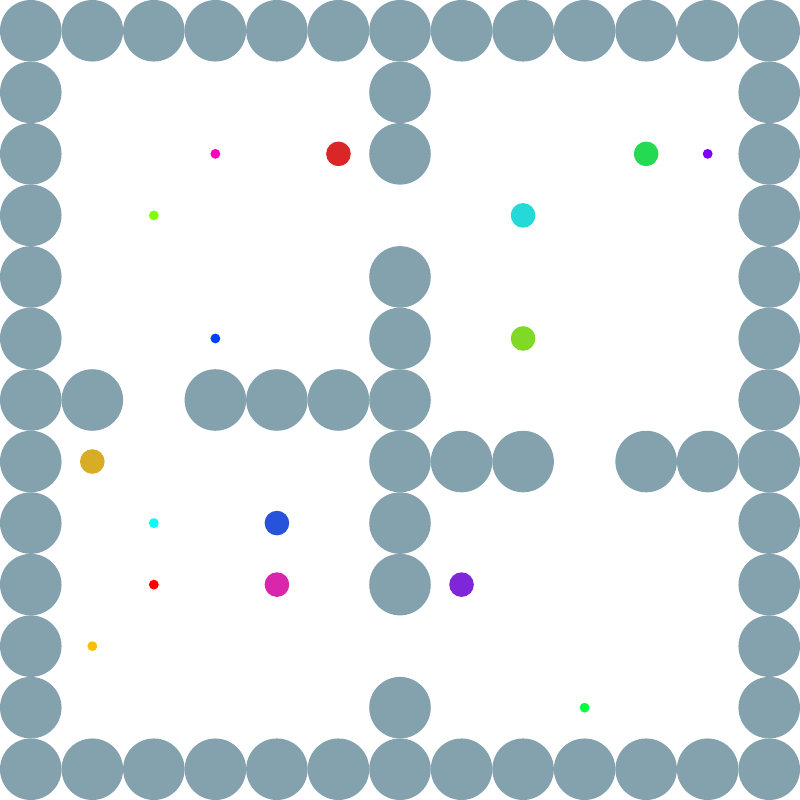}
        \caption{\texttt{string\_grid}}
        \label{subfig:maps_string_grid}
    \end{subfigure}
    \begin{subfigure}{85pt}
        \centering
        \includegraphics[width=80pt]{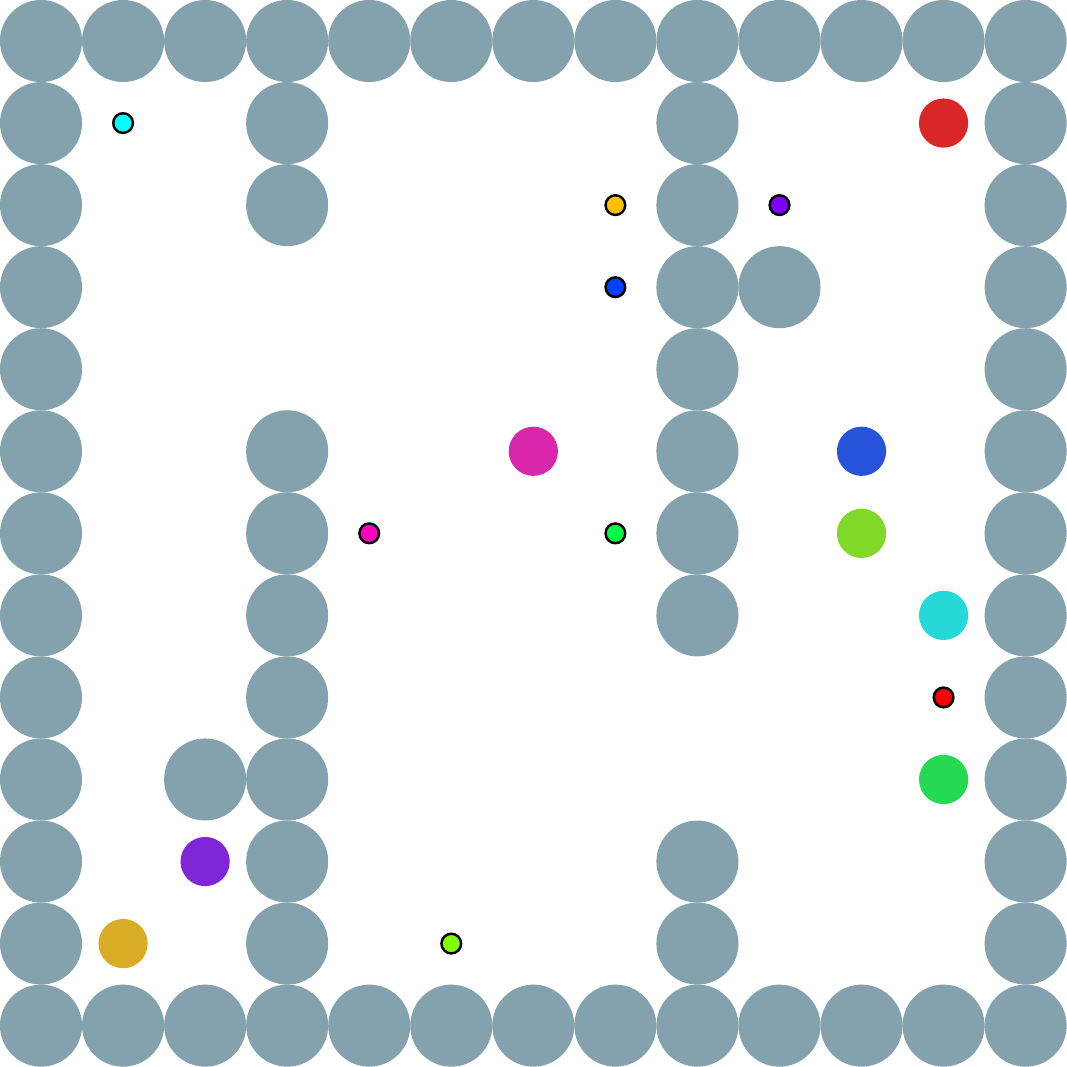}
        \caption{batched\_string\_grid}
        \label{subfig:maps_batched_string_grid}
    \end{subfigure}
    \caption{A rich collection of maps for multi-agent planning in continuous spaces in CAMAR: support for both continuous and grid landscapes together with MovingAI collection~\cite{sturtevant2012benchmarks}.}
    \label{fig:maps}
    \vspace{-5px}
\end{figure*}

Although every object in CAMAR is represented as a circle, it is still possible to create complex and detailed maps. A complex structure can be made by combining many smaller circles close together. The more circles that are used, the more accurate the map becomes. This method allows users to simulate walls, tunnels, mazes, and other complicated shapes even though the basic element is always a circle.

CAMAR includes several types of built-in maps. It also gives users the ability to add custom map generators, even if they are not compatible with JAX just-in-time compilation ~\cite{jax2018github}. A custom map generator can be connected easily by using the \texttt{string\_grid} or batched\_string\_grid formats. This flexibility allows users to design many kinds of environments, from simple random setups to complex and realistic maps.

The current set of built-in maps and generators includes (see Fig.~\ref{fig:maps}):
\begin{itemize}
    \item[\ref{subfig:maps_random_grid}] \texttt{random\_grid:} A map where obstacles, agents, and goals are placed randomly on a grid with a predefined size.
    \item[\ref{subfig:maps_labmaze_grid}] \texttt{labmaze\_grid:} Maps generated using LabMaze\footnote{https://github.com/google-deepmind/labmaze}~\cite{beattie2016deepmind} - maze generator with connected rooms.
    \item[\ref{subfig:maps_movingai}] \texttt{movingai:} Integrated two-dimensional maps from the MovingAI benchmark~\cite{sturtevant2012benchmarks}, adapted for continuous planning tasks. They can also be used in a \texttt{batched\_string\_grid} manner.
    \item[\ref{subfig:maps_caves_cont}] \texttt{caves\_cont:} A continuous type of map where caves are generated using Perlin noise, a common method in video games for creating realistic and varied landscapes.
    \item[\ref{subfig:maps_string_grid}] \texttt{string\_grid:} A grid map based on a text layout. Obstacles are placed according to characters in a string, similar to the MovingAI benchmark~\cite{sturtevant2012benchmarks}. Agent and goal positions can be fixed or random, depending on the free cells in the string.
    \item[\ref{subfig:maps_batched_string_grid}] \texttt{batched\_string\_grid:} Similar to \texttt{string\_grid}, but supports different obstacle layouts across parallel environments. This allows training on multiple map variations at once.
\end{itemize}

\paragraph{Reward Function}
CAMAR uses a scalar reward for each agent at every time step. This reward is the sum of four terms: a goal reward, a collision penalty, a movement-based reward, and a collective success reward:

\begin{equation*}\label{eq:reward}
r_i(t) = r_{\text{all\_g}}(t) + r_{\text{on\_g}_i}(t) + r_{\text{collision}_i}(t) + r_{\text{g\_dist}_i}(t)    
\end{equation*}

The terms are defined as follows:
\begin{eqnarray*}\label{eq:reward_det}
\begin{cases} 
r_{\text{all\_g}}(t) = + 0.5, \ \text{if } \forall i: \vec{x}_i(t) - \vec{x}_{\text{g}_i} \| \leq R_{\text{g}};\\
r_{\text{on\_g}_i}(t) = + 0.5, \ \text{if } \| \vec{x}_i(t) - \vec{x}_{\text{g}_i} \| \leq R_{\text{g}}; \\
r_{\text{collision}_i}(t) = - 1, \ \text{if } \exists j: \| \Delta \vec{x}_{ij}(t) \| < d_{\min}; \\
r_{\text{g\_dist}_i}(t) = + \texttt{shaping} \cdot \\ \hfill \cdot \left(\| \vec{x}_i(t-dt) - \vec{x}_{\text{g}_i} \| - \| \vec{x}_i(t) - \vec{x}_{\text{g}_i} \|\right)
\end{cases}
\end{eqnarray*}

Here, $\vec{x}_i(t)$ is the position of agent $i$ at time $t$, $\vec{x}_{\text{g}_i}$ is the goal position for agent $i$, $R_{\text{g}}$ is the distance threshold to count as reaching the goal (goal radius), $\Delta \vec{x}_{ij}(t)$ is the vector between agent $i$ and another object $j$, $d_{\min}$ is the minimal distance between agent $i$ and object $j$ (for circle objects $d_{\min} = R_i + R_j$ where $R_i$ and $R_j$ are radii), \texttt{shaping} is a user-defined coefficient that controls the strength of the movement-based term.

To support cooperation, the environment gives an extra reward when all agents reach their goals. In this case, each agent receives an additional reward of +0.5.

\subparagraph{Heterogeneous agents}

\begin{figure}[htb!]
    \centering
    \includegraphics[width=0.9\linewidth]{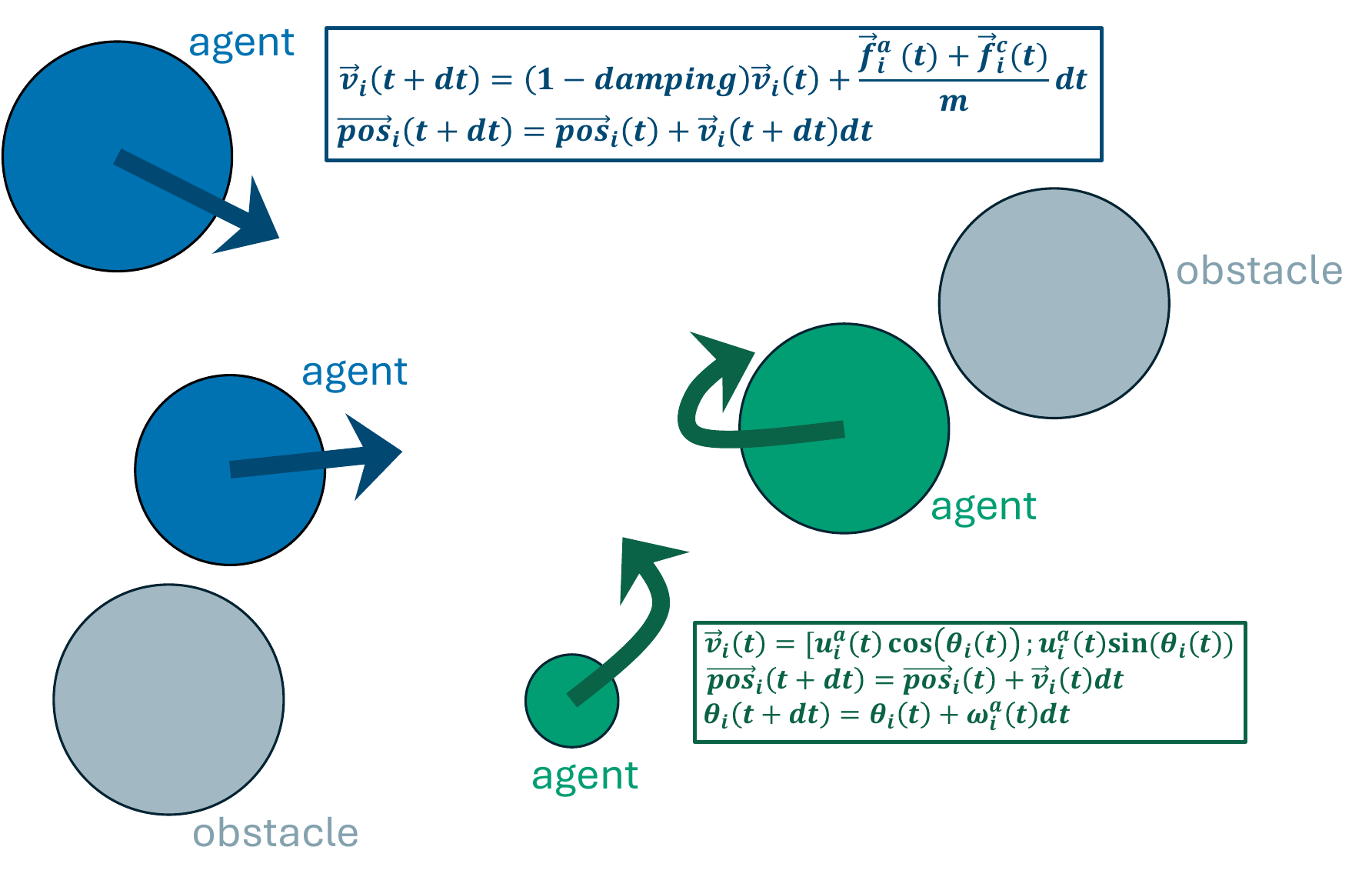}
    \caption{Illustration of heterogeneous agents with different sizes and dynamics supported by CAMAR. Blue agents are governed by \texttt{HolonomicDynamic}, while green agents follow \texttt{DiffDriveDynamic}. All agents navigate a shared environment while avoiding gray obstacles.}
    \label{fig:dynamics}
\end{figure}

Additionally, CAMAR supports heterogeneous agents in both size and dynamics. All map generators can produce agents with different properties, making it possible to study diverse multi-agent systems inspired by real-world scenarios.

For example, some agents can use \texttt{HolonomicDynamic}, while others follow \texttt{DiffDriveDynamic}. These agents operate together in a shared space, interact with the same obstacles, and must coordinate their movements despite having different control rules (Fig.~\ref{fig:dynamics}). Each agent follows its own dynamics model, but all agents contribute to a single global simulation.

CAMAR also supports agents with different sizes. Each agent can have its own radius, which affects how it moves and avoids collisions. This adds complexity to coordination and planning.

\paragraph{Metrics}

The evaluation metrics are defined as follows: the success rate is $\mathrm{SR} = \frac{1}{N} \sum_{i=1}^{N} \mathbf{1} \{ \| \vec{x}_i(T) - \vec{x}_{\text{goal}} \| \leq R_{\text{g}} \}$; the flowtime is $\mathrm{FT} = \frac{1}{N} \sum_{i=1}^{N} t_i$; the makespan is $\mathrm{MS} = \max_{i=1,\dots,N} t_i$; and the coordination is $\mathrm{CO} = 1 - \frac{C}{N \times T}$. Here, $N$ is the number of agents, $T$ is the maximum episode length, $t_i$ is the time step when agent $i$ first reaches its goal, $t_i = T$ for unfinished agents, $C$ is the total per-agent-per-timestep number of collisions over the episode.

\paragraph{Evaluation Protocols}

To support rigorous and reproducible benchmarking, CAMAR includes a standardized suite of evaluation protocols inspired by and extending prior work on cooperative MARL evaluation \cite{gorsane2022towards}. We adapt that framework for continuous multi-agent pathfinding, focusing on generalization across both agent count and map structure.

\begin{table*}[hbt!]
\centering
\small
\begin{tabular}{lcccccccc}
\toprule
 & \multicolumn{4}{c}{\texttt{random\_grid}} & \multicolumn{4}{c}{\texttt{labmaze\_grid}} \\
\cmidrule(lr){2-5} \cmidrule(lr){6-9}
Algorithm & SR $\uparrow$ & FT $\downarrow$ & MS $\downarrow$ & CO $\uparrow$
& SR $\uparrow$ & FT $\downarrow$ & MS $\downarrow$ & CO $\uparrow$ \\
\midrule
IPPO	 & 0.410±0.001 &  1695±10 &  160.0±0.0 &  1.000±0.000 & 0.213±0.013 &  2104±14 &  160.0±0.0 &  1.000±0.000 \\
MAPPO	 & \cellcolor{best} 0.830±0.001 &  984±5 &  151.4±0.3 &  1.000±0.000 & 0.568±0.004 &  1484±8 &  160.0±0.0 &  1.000±0.000 \\
IDDPG	 & 0.335±0.001 &  1851±10 &  160.0±0.0 &  1.000±0.000 & 0.167±0.000 &  2772±14 &  160.0±0.0 &  0.996±0.000 \\
MADDPG	 & 0.041±0.000 &  2508±12 &  160.0±0.0 &  0.913±0.001 & 0.027±0.000 &  2745±12 &  160.0±0.0 &  0.854±0.001 \\
ISAC	 & 0.115±0.001 &  2523±14 &  160.0±0.0 &  1.000±0.000 & 0.047±0.000 &  2808±12 &  160.0±0.0 &  1.000±0.000 \\
MASAC	 & 0.281±0.001 &  1843±11 &  160.0±0.0 &  0.856±0.001 & 0.105±0.001 &  2098±12 &  160.0±0.0 &  0.781±0.001 \\
\midrule
RRT*+IPPO	 & 0.420±0.001 &  1426±9 &  160.0±0.0 &  1.000±0.000 & 0.511±0.001 &\cellcolor{best}1316±6 &  160.0±0.0 &  0.999±0.000 \\
RRT*+MAPPO	 & 0.828±0.001 & \cellcolor{best} 971±5 & \cellcolor{best} 150.4±0.3 &  1.000±0.000 & 0.556±0.001 & 1326±7 &  160.0±0.0 &  0.999±0.000 \\
RRT*+IDDPG	 & 0.280±0.001 &  2181±12 &  160.0±0.0 &  1.000±0.000 & 0.189±0.000 &  2635±14 &  160.0±0.0 &  0.997±0.000 \\
RRT*+MADDPG	 & 0.037±0.000 &  2953±15 &  160.1±0.0 &  0.984±0.000 & 0.037±0.000 &  2918±14 &  160.1±0.0 &  0.969±0.000 \\
RRT*+ISAC	 & 0.143±0.000 &  2618±13 &  160.0±0.0 &  1.000±0.000 & 0.058±0.000 &  2749±13 &  160.0±0.0 &  1.000±0.000 \\
RRT*+MASAC	 & 0.054±0.000 &  2511±14 &  160.0±0.0 &  1.000±0.000 & 0.034±0.000 &  2854±15 &  160.0±0.0 &  0.994±0.000 \\
\midrule
RRT*+PD	 & 0.678±0.002 &  2010±59 &  160.0±0.0 &  0.997±0.000 & \cellcolor{best} 0.692±0.004 &  1807±49 &  160.0±0.0 &  0.971±0.002 \\
RRT+PD	 & 0.413±0.014 &  2440±264 &  160.0±0.0 &  0.788±0.021 & 0.528±0.021 &  2049±251 &  160.0±0.0 &  0.558±0.025 \\
\bottomrule
\end{tabular}
\caption{Extended performance comparison across different algorithms in the \texttt{random\_grid} and \texttt{labmaze\_grid} environments including integrated RRT* with off-policy algorithms additionally. 
Reported metrics are SR (Success Rate), FT (Flowtime), MS (Makespan), and CO (Coordination), each shown as IQM±$\text{CI}_{95}$. Confidence intervals are symmetric for clarity and computed using 1K bootstrapped samples. Arrows indicate the direction of improvement: $\uparrow$ denotes higher is better, $\downarrow$ indicates lower is better. \compactbest{Tan boxes} highlight the best-performing approach. The CO metric is not colored here for visibility.} 
\label{tab:performance_comparison}
\end{table*}

We propose three evaluation tiers: \textbf{Easy}, \textbf{Medium}, and \textbf{Hard}, each targeting a different level of generalization. \textbf{Easy} evaluates performance on unseen start and goal positions using the same map type and number of agents as during training. \textbf{Medium} tests generalization to maps with similar structure but a different number of agents and obstacle parameters. \textbf{Hard} measures generalization to fully unseen map types from the MovingAI street collection, often with a different number of agents than during training.

Each tier follows a defined training and evaluation setup using introduced metrics, aggregated by the Interquartile Mean (IQM) with 95\% confidence intervals ($\text{CI}_{95}$) for fair comparison across methods and difficulty levels.

All experiments use fixed JAX ~\cite{jax2018github} random seeds for reproducibility. In total, each protocol involves training multiple models and running thousands of evaluation episodes. We include detailed training and evaluation scripts and provide all protocol maps in the public CAMAR repository.

These protocols help the community better track progress on continuous multi-agent pathfinding and identify which algorithms generalize well to more realistic conditions. Sample efficiency curves, metric-vs-agent-count plots, and performance profiles are supported and recommended for deeper analysis.

\section{Experimental Evaluation}

In this section, we evaluate both the scalability and benchmarking capabilities of our environment.
We begin by training and testing a set of popular  MARL algorithms, as well as classical non-learnable and hybrid methods. These experiments show that the environment supports a wide range of navigation and coordination strategies. We also present results from a simple heterogeneous-agent scenario to demonstrate support for heterogeneous MARL research. Finally, we measure the performance of the simulator in terms of simulation speed, and compare it with VMAS using a shared experimental setup.

\subsection{Experimental Setup}

We evaluate 6 MARL algorithms: IPPO, MAPPO, IDDPG, MADDPG, ISAC~\cite{haarnoja2018soft}, and MASAC. In addition, we include 2 non-learnable baselines, RRT+PD and RRT*+PD, and 6 hybrid methods: RRT*+IPPO, RRT*+MAPPO, RRT*+IDDPG, RRT+MADDPG, RRT*+ISAC, and RRT*+MASAC.

All methods are evaluated on two procedurally generated map types: \texttt{random\_grid} and \texttt{labmaze\_grid}, each with 6 versions that vary in obstacle density and agent count (8 or 32)\footnote{Our current Medium-tier protocol includes tasks with 8 and 32 agents, but we plan to extend it in future versions to support larger agent populations and more complex settings as methods advance.}. For \texttt{labmaze\_grid}, an additional connection probability ranging from 0.4 to 1.0 is used to test different maze complexities. Generation details for all training and evaluation maps are provided in Appendix D.

The “independent” variants (IPPO~\cite{de2020independent}, IDDPG, ISAC) train each agent using its own policy and value function. These methods do not use centralized critics or information sharing across agents. In contrast, the multi-agent versions (MAPPO, MADDPG, MASAC) use centralized critics during training to improve coordination. All approaches use parameter sharing, meaning that agents use the same neural network weights.

Each algorithm is trained for 20M (IPPO, MAPPO) and 2M (IDDPG, MADDPG, ISAC, MASAC) steps per scenario. The training is done independently for each of the 12 map variations. After training, we evaluate the models on both seen and unseen tasks to test their generalization. In total, we train 532 models and evaluate them across 5184 tasks, with 1000 episodes per task. The experiments were run on a single NVIDIA H100 GPU and took around 1000 hours in total.

For the non-learning baselines, we use RRT+PD and RRT*+PD. These methods use classical planning algorithms to generate a path to the goal for each agent. Each path is generated using either RRT (with 50,000 iterations)~\cite{lavalle1998rapidly} or RRT* (with 3000 iterations). The agent then follows the path using a simple PD controller.

We also evaluate hybrid methods where agents receive additional RRT* information during training and evaluation. At the start of each episode, RRT* generates sample paths from the goal to the agent’s position. These paths and their estimated costs are included in the agent’s observation, enabling the policy to learn from approximate cost-to-go values without invoking RRT* at every step.

To evaluate simulator performance, we measure simulation speed in steps per second (SPS) on a $20 \times 20$ \texttt{random\_grid} map with 0.3 obstacle density (120 obstacles). We vary the number of agents and parallel environments to assess CAMAR’s scalability. For fair comparison, we benchmark CAMAR against VMAS~\cite{bettini2022vmas} using identical map size, agent count, and a single NVIDIA H100 GPU.

\subsection{Benchmark}

The main results are shown in Table~\ref{tab:performance_comparison}. On the \texttt{random\_grid} map, MAPPO reaches the highest success rate with strong coordination. Adding planning improves efficiency. RRT*+MAPPO gives faster routes than MAPPO, and RRT*+IPPO improves over IPPO. The classic RRT*+PD baseline reaches high success without learning, but shows low coordination because it plans for each agent alone.

For off-policy algorithms, results are mixed. RRT* improves ISAC and IDDPG slightly, but weakens MASAC and MADDPG. These methods use a centralized critic and must process long input vectors that include RRT* features. This makes training unstable and harms generalization. Among MARL baselines, IPPO and IDDPG reach good success but slower flowtime than MAPPO. MASAC and MADDPG fail in both maps.

The \texttt{labmaze\_grid} setting is harder due to narrow corridors and sparse rewards. All MARL baselines drop in success. RRT*+PD performs best, which shows the value of full-path planning when learning signals are weak. The simpler RRT+PD planner gives worse paths but still beats many MARL baselines in success.

More detailed analysis of these results and hybrid methods is given in Appendix D.

\subsection{Heterogeneous Agents}

To demonstrate support for heterogeneous teams, we extend the \texttt{give\_way} task so that two agents must pass through a narrow corridor where only the smaller red agent can enter the central chamber. The larger blue agent must wait.

\begin{figure}[htb!]
    \vspace{-3px}
    \centering
    \begin{subfigure}{140pt}
        \centering
        \includegraphics[width=140pt]{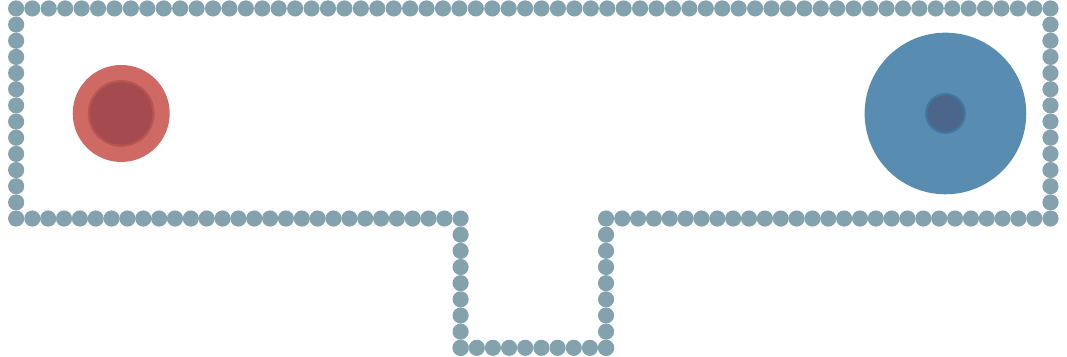}
        \caption{\texttt{hetero\_give\_way} scenario}
        \label{subfig:hetero_vis}
    \end{subfigure}
    \hspace{10pt}
    \begin{subfigure}{80pt}
        \centering
        \footnotesize
        \begin{tabular}{lc}
        Algorithm & SR \\
        \midrule
        IPPO & 0.5 \\
        HetIPPO & 0.7 \\
        MAPPO & 0.5 \\
        HetMAPPO & 0.0\\
        \bottomrule
        \end{tabular}
        \caption{SR results.}
        \label{subfig:hetero_res}
    \end{subfigure}
    \label{fig:hetero_give_way}
\caption{Example of heterogeneous agent coordination (a). Success rates of algorithms are shown in (b).}

\end{figure}

We compare IPPO and MAPPO with shared policies to their heterogeneous versions, where each agent has its own model. As shown in Fig.~\ref{subfig:hetero_res}, HetIPPO performs better, but HetMAPPO fails, likely because the centralized critic cannot handle the larger and more diverse input space.

This experiment shows that CAMAR can model agents with different sizes and abilities and is suitable for studying coordination in heterogeneous teams. More details are in Appendix D.

\subsection{Scalability Analysis}

We study how CAMAR scales when we increase the number of parallel environments, agents, and obstacles. The full results are shown in Fig. \ref{fig:scalability_analysis}. Visual examples of CAMAR and VMAS running the same scenario are provided in the appendix E.

We first fix the number of agents at 32 and measure the simulation speed when we increase the number of parallel environments from \num{5} to \num{6000}+. VMAS scales roughly linearly but stays below \num{10000} steps per second (SPS) at \num{6000} environments. In contrast, CAMAR rises quickly to about \num{1000} environments and then stays close to \num{50000} SPS even as we add more parallel environments. This shows that CAMAR can support fast and stable training with many vectorized environments.

Next, we fix the number of environments at \num{2000} and increase the number of agents from \num{4} to \num{128}. CAMAR keeps more than \num{100000} SPS when the agent count is below \num{16} and remains above \num{10000} SPS even at \num{128} agents. VMAS begins near \num{20000} SPS with \num{4} agents but drops to about \num{500} SPS at \num{128} agents. In this setting, CAMAR is up to \num{20} times faster when many agents act together.

\begin{figure}[tb!]
    \centering
    \begin{subfigure}[b]{\linewidth}
        \includegraphics[width=\linewidth]{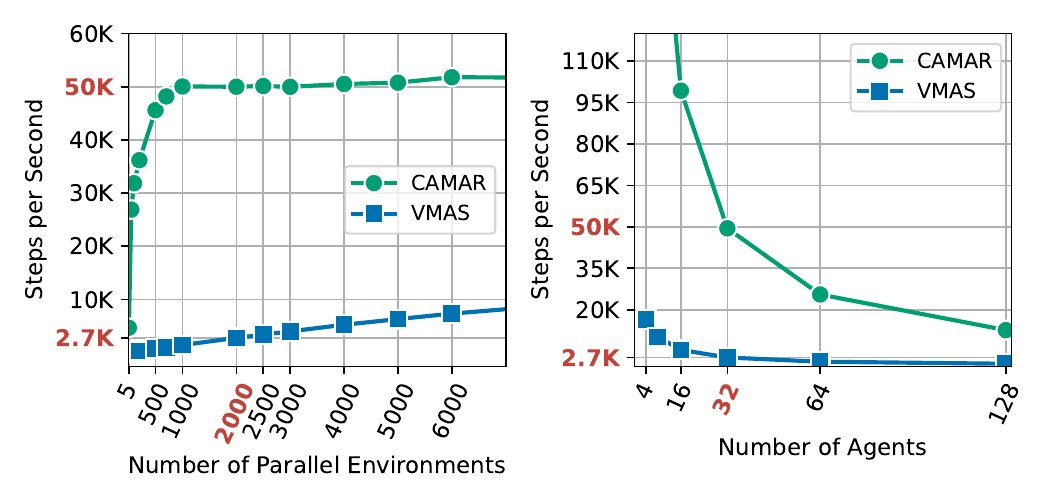}
        \caption{CAMAR vs VMAS with increasing number of environments, agents, and obstacles.}
        \label{subfig:scalability_analysis_a}
    \end{subfigure}%
    \hspace{0.01pt}
    \begin{subfigure}[b]{0.981\linewidth}
        \includegraphics[width=\linewidth]{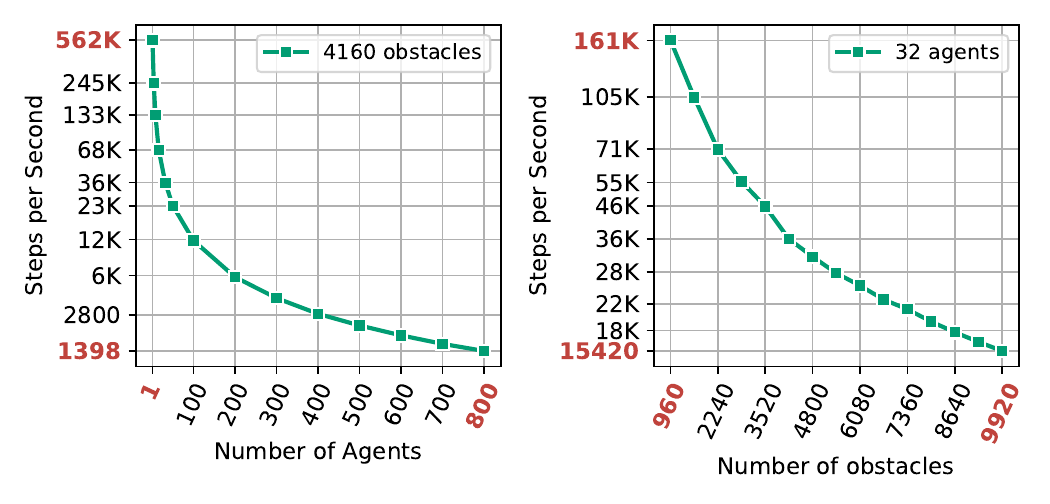}
        \caption{CAMAR handles up to 800 agents, maintaining 1400 SPS and high observation throughput.}
        \label{subfig:scalability_analysis_b}
    \end{subfigure}
    \caption{CAMAR achieves robust scalability in multi-agent simulation, surpassing VMAS in performance across varied settings and supporting efficient simulation of large agent populations with high observation rates.}
    \label{fig:scalability_analysis}
\end{figure}

We also test extreme conditions by keeping \num{4160} fixed and increasing the number of agents up to \num{800}. The speed goes down as more agents are added, but CAMAR still maintains about \num{1400} SPS at \num{800} agents. Since every agent produces one observation per step, the total amount of data remains high.

Finally, we test how obstacle count alone affects speed. We fix the number of agents at \num{32} and increase the number of obstacles up to \num{9920}. CAMAR still reaches about \num{15000} SPS in this most cluttered setting. This shows that the system remains robust even in very dense maps.

In summary, CAMAR achieves around \num{50000} SPS on complex scenarios with many obstacles and \num{32} agents. The speed depends mainly on the number of objects in the scene. In other tasks it reaches up to \num{161000} SPS with \num{32} agents and \num{960} obstacles, and up to \num{562000} SPS when simulating a single agent with \num{4160} obstacles. These results confirm that CAMAR can operate above \num{100000} SPS.
Visualisations of the test scenes are shown in the appendix E. Further comparisons with other MARL environments are also given in the appendix E.

\section{Acknowledgments}

The study was supported by the Ministry of Economic Development of the Russian Federation (agreement No. 139-15-2025-013, dated June 20, 2025, IGK 000000C313925P4B0002).

\section{Conclusion}

This paper introduces CAMAR, a high-performance benchmark for continuous-space multi-agent reinforcement learning. CAMAR combines realistic dynamics with efficient simulation, supporting over \num{100000} steps per second using JAX~\cite{jax2018github}. It includes a diverse set of navigation tasks, a standardized evaluation protocol with built-in metrics, and a range of strong baselines from both classical, learning-based, and hybrid methods. These components enable reliable, scalable, and reproducible evaluation of MARL algorithms.

\bibliographystyle{unsrt}
\bibliography{bib}


\appendix

\section*{Appendix Contents}

\begin{table}[ht]
    \centering
    \Large
    \begin{tabular}{ll}
    \textbf{Sections} & \textbf{Contents} \\ \toprule
    Appendix A & \begin{tabular}[c]{@{}l@{}}Evaluation Protocol\end{tabular} \\ \midrule
    Appendix B & \begin{tabular}[c]{@{}l@{}}Code Examples\end{tabular} \\ \midrule
    Appendix C & \begin{tabular}[c]{@{}l@{}}Discussion of Limitations\end{tabular} \\ \midrule
    Appendix D & \begin{tabular}[c]{@{}l@{}}Extended Benchmark Results\end{tabular} \\ \midrule
    Appendix E & \begin{tabular}[c]{@{}l@{}}Scalability Analysis Details\end{tabular} \\ \midrule
    Appendix F & \begin{tabular}[c]{@{}l@{}}Future Work and Directions\end{tabular} \\ \midrule
    Appendix G & \begin{tabular}[c]{@{}l@{}}Implementation Details\end{tabular} \\ \midrule
    Appendix H & \begin{tabular}[c]{@{}l@{}}Extended Related Work\end{tabular} \\ \midrule
    \end{tabular}
\vspace{-1em}
\end{table}

\section{Appendix A --- Evaluation Protocol}
\label{app:evaluation}

\paragraph{Evaluation Suite Overview.}

To make results easy to compare and fully reproducible, CAMAR adopts the “Standardised Performance Evaluation Protocol for Cooperative MARL” by \cite{gorsane2022towards} and \emph{extends} it to continuous multi-agent path-finding.  
We keep the core ideas, fixed training budgets, multiple random seeds, and strict uncertainty estimates, but add path-finding-specific stress-tests:

\begin{itemize}
    \item \textbf{Axis 1: Agent count.} We vary the number of agents to see whether a method still works when the team grows.
    \item \textbf{Axis 2: Map difficulty.} We change obstacle density and map geometry to check if the learned policy still solves harder layouts.
\end{itemize}

\paragraph{Three difficulty tiers.}

\begin{itemize}
    \item \textbf{Easy} uses the same map, agent count, and obstacle parameters for training and testing. Only the random seeds that place starts and goals change. This tells us whether an algorithm can \emph{solve} the problem at all.
    

    \item \textbf{Medium} trains on one map setting but evaluates on 12 variants that share the same map types while changing agent count and obstacle density. This probes generalization within the same domain and includes testing generalization across agents. The current protocol is designed for reproducibility and fair comparison across methods. However, since the environment scales efficiently, we plan to extend the Medium tier in future releases to include a wider range of agent counts and task complexities as MARL methods progress.
    
    \item \textbf{Hard} trains on any user-chosen maps—except the MovingAI street set—and then tests on those unseen street maps with new agent counts. This is a near-real-world stress test.
\end{itemize}

\begin{mybox}{Standardized Evaluation Suite for CAMAR}

\footnotesize
\textbf{Input.}  A set of maps $\mathcal{M}$, task sets $\mathcal{T}_m$, and a pool of algorithms $\mathcal{A}$.

\textbf{1.\ Default settings}
\begin{itemize}
    \item Training steps $T$: \textbf{2M} (off-policy) or \textbf{20M} (on-policy).
    \item Independent runs $R$: \textbf{3} seeds.
    \item Evaluation episodes $E$: \textbf{1\,000} per interval.
    \item Evaluation intervals $\mathcal{I}$ (based on available resources): every \textbf{10K-100K} steps (off-policy) or \textbf{100K-1000K} steps (on-policy).
\end{itemize}

\textbf{2.\ Metrics}
\begin{itemize}
    \item \emph{Returns} $G$ for sample-efficiency plots.
    \item \emph{Success Rate} (SR), \emph{Flowtime} (FT), \emph{Makespan} (MS), and \emph{Coordination} (CO).
    \item Per-task: mean $G^a_t$ over $E$ episodes with 95\% CIs.
    \item Per-protocol: build an $(R{\times}|\mathcal{T}|)$ matrix of normalised returns, then report IQM, optimality gap, probability of improvement, and performance profiles (all with 95\% bootstrap CIs).
\end{itemize}

\textbf{3.\ Three difficulty tiers}
\begin{enumerate}
    \item \textbf{Easy}: train and test on the same map/agent count.  
          12 models $\times$ 3 seeds.
    \item \textbf{Medium}: train on one map, test on all 12 maps and aggregate.  
          12 models $\times$ 3 seeds.
    \item \textbf{Hard}: train on any maps (except MovingAI), test on MovingAI~\cite{sturtevant2012benchmarks} street maps with new agent counts.  
          Single model (preferably 3 seeds).
\end{enumerate}

\textbf{4.\ Reporting checklist}
\begin{itemize}
    \item Hyper-parameters, network sizes, and compute budget.
    \item Map generation settings for each tier.
    \item Final IQM $\pm$95\% CI for SR, FT, MS, CO (mandatory).
          Sample-efficiency curves are recommended for Easy and Medium, optional for Hard.  
          In Hard also plot each metric against the number of agents to test the ability to scale.
\end{itemize}
\label{mybox:protocol_text}
\end{mybox}

\begin{table*}[htb!]
\small
\centering
\begin{tabular}{p{70pt}p{125pt}p{125pt}p{125pt}}
\toprule
 & \textbf{Easy} & \textbf{Medium} & \textbf{Hard} \\
\midrule
\textbf{Purpose} & Test that the method can solve the problem without testing generalization. & Test that the method can solve the problem and generalise across similar map types including varying number of agents. & Test that the method can generalise to near-real-world settings. \\
\midrule
\textbf{How to train and evaluate} & \textbf{Train on} \texttt{map\_X} (see Appendix). & \textbf{Train on} \texttt{map\_X} (see Appendix). & \textbf{Train on} any map collection excluding MovingAI \\
 & \textbf{Evaluate on} \textit{the same} \texttt{map\_X}. & \textbf{Evaluate on} \textit{all 12} maps. & \textbf{Evaluate on} MovingAI street collection with varying agent counts \\
 & \textbf{Repeat} for all 12 maps & \textbf{Repeat} for all 12 maps & \\
\midrule
\textbf{Number of models} & 12 trained models $\times$ 3 seeds & 12 trained models $\times$ 3 seeds & 1 trained model \\
\textbf{Number of evals} & 12 evaluations $\times$ 1K episodes & 144 evaluations $\times$ 1K episodes & 30 evaluations $\times$ 1K episodes $\times$ varying number of agents \\
\textbf{Report} & \textbf{Metrics} - if needed & \textbf{Metrics} - mandatory & \textbf{Metrics} - mandatory\\
 & \textbf{sample-efficiency curves} - strongly recommended & \textbf{sample-efficiency curves} - if needed & \textbf{sample-efficiency curves} - if resources allow \\
 & & & \textbf{metrics vs num\_agents} - mandatory \\
\bottomrule
\end{tabular}
\caption{Overview on 3-tier evaluation protocols presenting the purpose, number of trained models and evaluations for each protocol}
\label{tab:protocol_overview}
\end{table*}

Each tier follows the default budget: 2M steps for off-policy and 20M steps for on-policy algorithms, three random seeds, and evaluations every 10K-100K or 100K-1000K steps. We keep the JAX seed fixed at \texttt{5} and split it to generate all evaluation keys, which makes every run exactly repeatable.

\paragraph{Reporting and analysis.}
We require final IQM\,$\pm$\,95\% CI of Success Rate, Flowtime, Makespan, and Coordination.
Sample-efficiency curves are strongly recommended for Easy and Medium and optional for Hard if resources allow.
For the Hard tier, we also ask for plots of each metric versus the number of agents because scalability is a key question.

\section{Appendix B --- Code Examples}
\label{app:code}

The CAMAR library provides a simple and flexible interface for building custom multi-agent pathfinding environments. Below we show two examples that demonstrate how to create environments using different map generators and agent dynamics.

\paragraph{Example 1} (Fig.~\ref{code:camar_example1}) shows how to create an environment with the \texttt{random\_grid} map. It uses the string-based API to pass all configuration options directly. This includes the number of agents, the size range for agents and goals, and the dynamic model to use. We set the dynamic to \texttt{HolonomicDynamic}, and also customize the observation window and the shaping factor. This example shows how to use heterogeneous sizes for agents and goals, which can help simulate more realistic scenarios.

\begin{figure}[htb!]
    \centering
    \scriptsize
    \begin{tcolorbox}[left=1.5ex,top=0ex,bottom=-.5ex,toprule=1pt,rightrule=1pt,bottomrule=1pt,leftrule=1pt,colframe=black!20!white,colback=black!5!white, arc=1ex]
\begin{lstlisting}[language=Python]
from camar import camar_v0

env1 = camar_v0(
    map_generator="random_grid",
    dynamic="HolonomicDynamic",
    pos_shaping_factor=2.0,
    window=0.25, # obs window
    max_obs=10, # max num of obj in obs
    map_kwargs={
        "num_agents": 16,
        "agent_rad_range": (0.01,  0.05 ),
        "goal_rad_range":  (0.001, 0.005),
    },
    dynamic_kwargs={
        "mass": 2.0,
    }
)
\end{lstlisting}
    \end{tcolorbox}
    \caption{Example 1. Creating a CAMAR environment using random grid maps and holonomic agents. The environment is loaded using the string-based API, which supports configuration via YAML or inline dictionary.}
    \label{code:camar_example1}
\end{figure}

\paragraph{Example 2} (Fig.~\ref{code:camar_example2}) shows how to use maps from the MovingAI benchmark~\cite{sturtevant2012benchmarks}. These maps are loaded by name using the map generator function movingai(). This example also shows how to use heterogeneous agent dynamics by combining different models (e.g. \texttt{DiffDriveDynamic} and \texttt{HolonomicDynamic}). This is done using the class-based API with MixedDynamic. The number of agents of each type is specified, and the total is passed to the environment. This allows testing how different types of agents can cooperate in complex environments.

\begin{figure}[htb!]
    \centering
    \scriptsize
    \begin{tcolorbox}[left=1.5ex,top=0ex,bottom=-.5ex,toprule=1pt,rightrule=1pt,bottomrule=1pt,leftrule=1pt,colframe=black!20!white,colback=black!5!white, arc=1ex]
\begin{lstlisting}[language=Python]
from camar import camar_v0
from camar.maps import movingai
from camar.dynamics import MixedDynamic, DiffDriveDynamic, HolonomicDynamic

dynamics_batch = [
    DiffDriveDynamic(mass=1.0),
    HolonomicDynamic(mass=10.0),
]
# 8 diffdrive + 24 holonomic = 32 total
num_agents_batch = [8, 24]
mixed_dynamic = MixedDynamic(
    dynamics_batch=dynamics_batch,
    num_agents_batch=num_agents_batch,
)

map_generator = movingai(
    map_names=["street/Denver_0_1024",
               "bg_maps/AR0072SR"],
    num_agents=sum(num_agents_batch),
)

env2 = camar_v0(
    map_generator=map_generator,
    dynamic=mixed_dynamic,
    pos_shaping_factor=2.0,
    window=0.25,
    max_obs=10,
)
\end{lstlisting}
    \end{tcolorbox}
    \caption{Example 2. Creating a CAMAR environment with MovingAI maps~\cite{sturtevant2012benchmarks} and mixed agent dynamics. This example demonstrates the class-based API for explicitly specifying heterogeneous agent behavior.}
    \label{code:camar_example2}
\end{figure}

These examples can be used as templates for building new scenarios in CAMAR. Users can adjust map parameters, agent settings, or dynamic models to suit their research needs. The CAMAR design supports quick changes to both environment structure and agent behavior using only a few lines of code.

\section{Appendix C --- Limitations}
\label{app:limitations}

CAMAR offers fast and flexible simulation, but there are still important gaps. First, all built-in scenarios use static obstacles. We do not yet provide maps with moving obstacles, so agents cannot train on fully dynamic scenes. While the engine supports different sizes for agents, goals, and landmarks, the built-in map generators do not include variations in landmark sizes. We believe that adding new maps with obstacles of different sizes and dynamic behavior is an important direction for future work. Fortunately, CAMAR’s modular design makes it easy to extend in this way.

Second, although we support two built-in dynamics models (\texttt{HolonomicDynamic} and \texttt{DiffDriveDynamic}), and allow mixing them to build heterogeneous teams, both use basic integration schemes: either explicit or non-explicit Euler. To achieve stable simulation with larger time steps (for example, dt = 0.1), a frameskip must be applied. Without frameskip, simulation becomes unstable with large dt values, and a smaller integration step like dt = 0.005 must be used together with frameskip = 20. Adding more stable integration methods, such as Runge-Kutta, could improve stability and allow efficient simulation with fewer steps. For example, using Runge-Kutta with dt = 0.05 and frameskip = 2 (for a total of 8 integration steps per forward pass) might offer a better trade-off between speed and accuracy.

Third, there is no built-in communication mechanism between agents. Adding message passing would help study how algorithms use shared information.

\section{Appendix D --- Extended Benchmark}
\label{app:benchmark}

\subsection{Performance Analysis}

\paragraph{Detailed Benchmark Analysis}

The main evaluation compares success rate (SR), flowtime (FT), makespan (MS), and coordination (CO) on two map types. On \texttt{random\_grid}, MAPPO gives the highest SR with strong CO. RRT*+MAPPO reaches similar SR but with lower FT and MS, which shows that planning improves movement efficiency. RRT*+IPPO gives smaller gains but still improves over IPPO. The classic RRT*+PD baseline reaches high SR but low CO because it plans for each agent alone and does not react to dynamic agent movement.

Hybrid methods with off-policy algorithms show mixed behavior. RRT* improves ISAC and IDDPG in SR and in some cases CO. But RRT* weakens MASAC and MADDPG, where SR is almost zero. These algorithms use a centralized critic with long input vectors, which include observations and RRT* samples from all agents. This leads to unstable training.

IPPO and IDDPG reach good SR but slower FT than MAPPO. ISAC gives high CO but low SR. MASAC and MADDPG fail to explore and reach very low SR. Additional experiments and analysis on DDPG and SAC can be found below.

The \texttt{labmaze\_grid} map is harder because of sparse rewards and narrow corridors. All MARL baselines drop in SR. RRT*+PD performs best in SR due to full-path planning, but CO is still low. The simple RRT+PD baseline gives worse paths and lower SR than RRT*+PD, but in some cases it still outperforms MARL baselines in SR.

These results show a clear trade-off. Planning gives strong reliability and better path efficiency. MARL gives better coordination but fails in sparse reward maps. Hybrid methods give only small gains. One reason is that the current policy networks are small and cannot process the large inputs that combine local observations with RRT* samples. Stronger model architectures may improve this integration.

\paragraph{Aggregate Analysis and Performance Profiles}

\begin{figure}[htb!]
    \centering
    \begin{subfigure}{0.49\linewidth}
        \includegraphics[width=\linewidth]{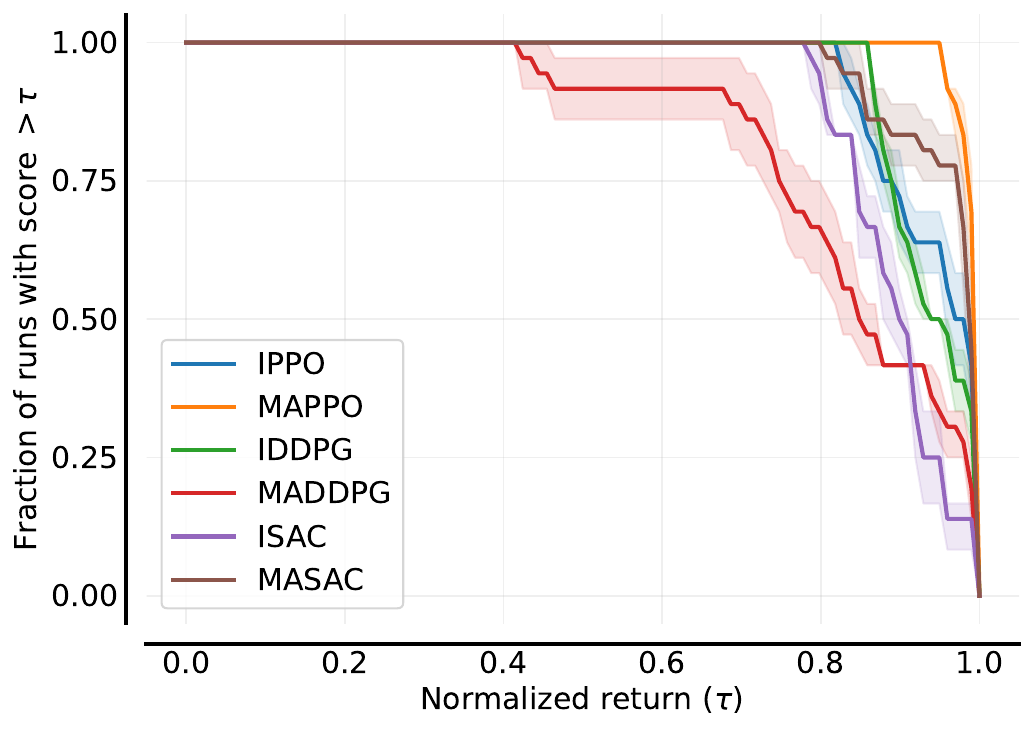}
    \end{subfigure}
    \begin{subfigure}{0.49\linewidth}
        \includegraphics[width=\linewidth]{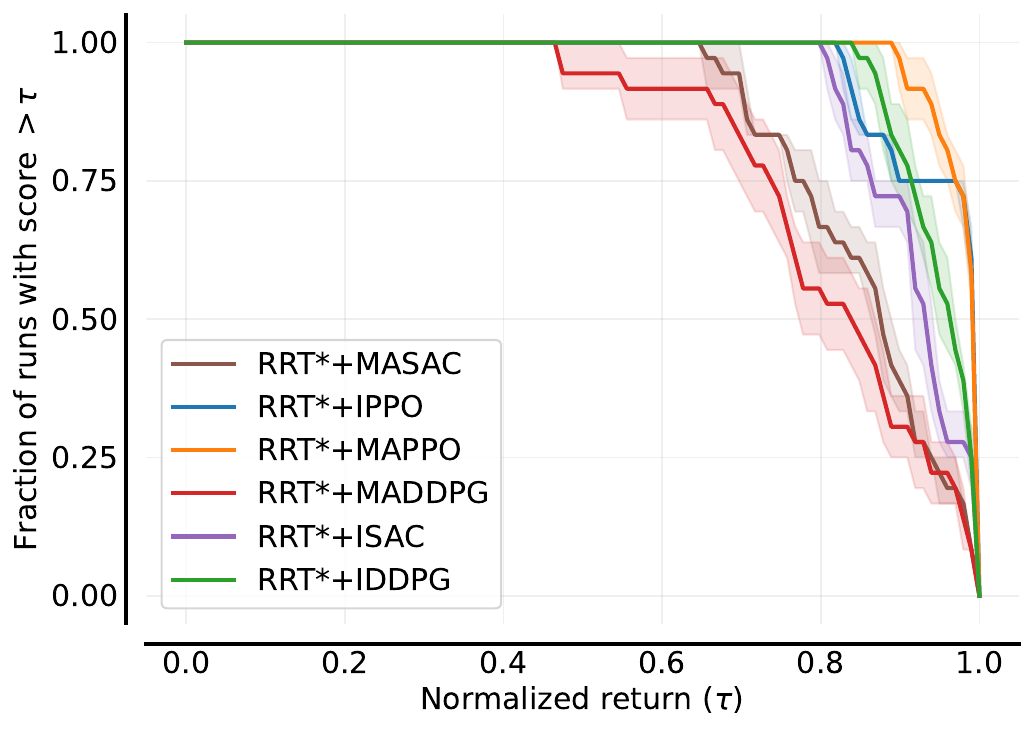}
    \end{subfigure}
    \caption{Performance profiles based on normalized return. Each line shows the fraction of evaluation runs that achieved a score above a given threshold $\tau$.}
    \label{fig:performance_profile_return}
\end{figure}

\begin{figure}[htb!]
    \centering
    \begin{subfigure}{0.49\linewidth}
        \includegraphics[width=\linewidth]{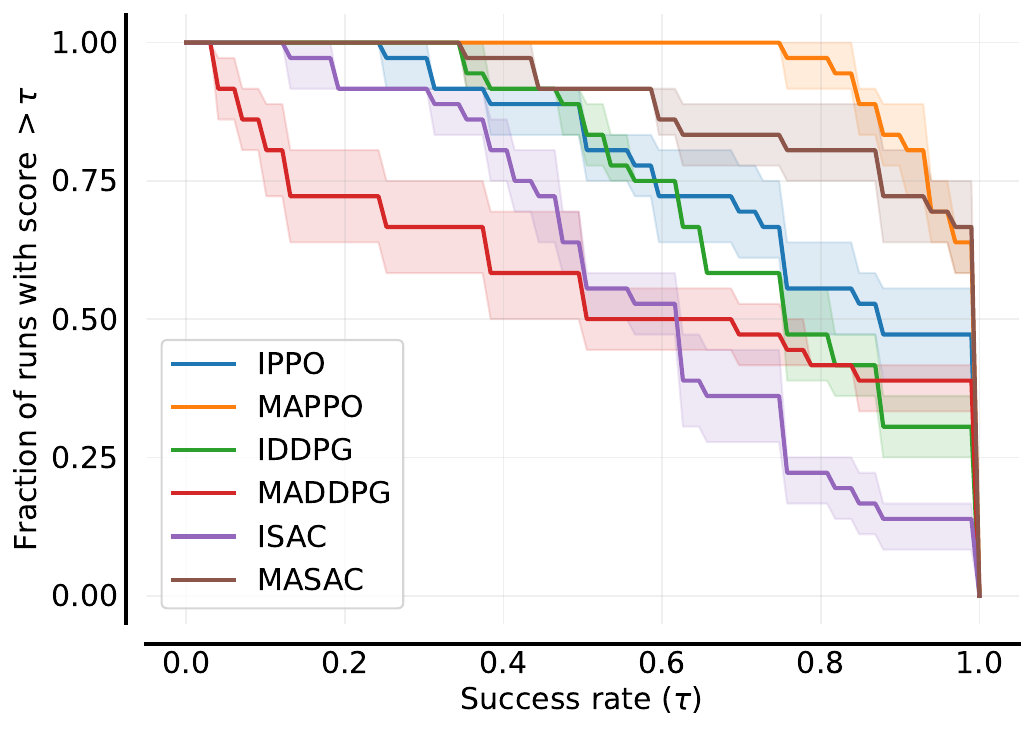}
    \end{subfigure}
    \begin{subfigure}{0.49\linewidth}
        \includegraphics[width=\linewidth]{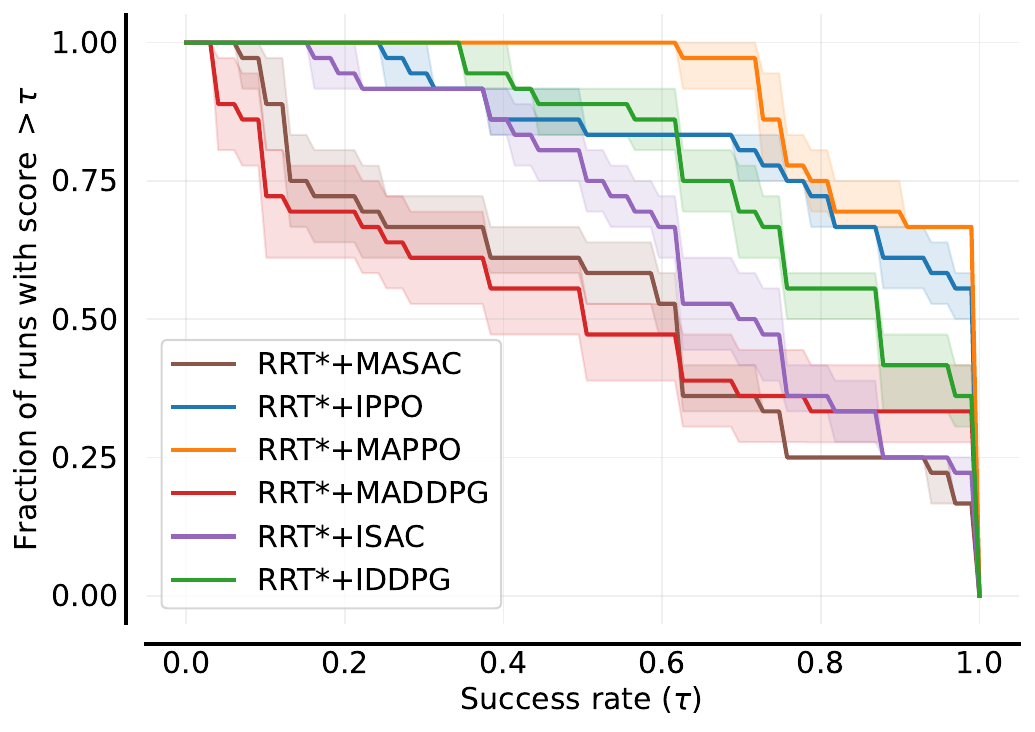}
    \end{subfigure}
    \caption{Performance profiles based on success rate. Each line shows the fraction of evaluation runs that achieved SR above a threshold $\tau$.}
    \label{fig:performance_profile_success_rate}
\end{figure}

\begin{figure*}[hbt!]
    \centering
    \begin{subfigure}{0.92\linewidth}
        \includegraphics[width=\linewidth]{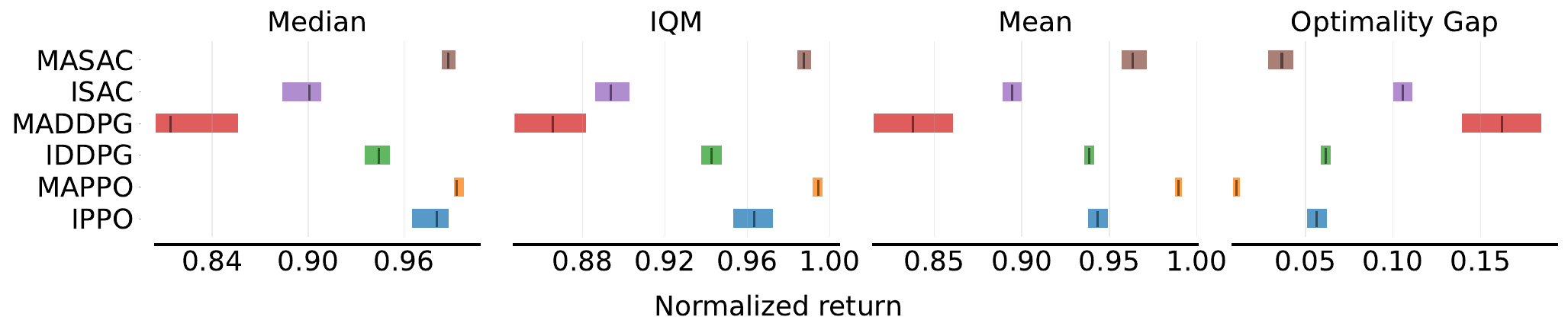}
    \end{subfigure}
    \begin{subfigure}{0.92\linewidth}
        \includegraphics[width=\linewidth]{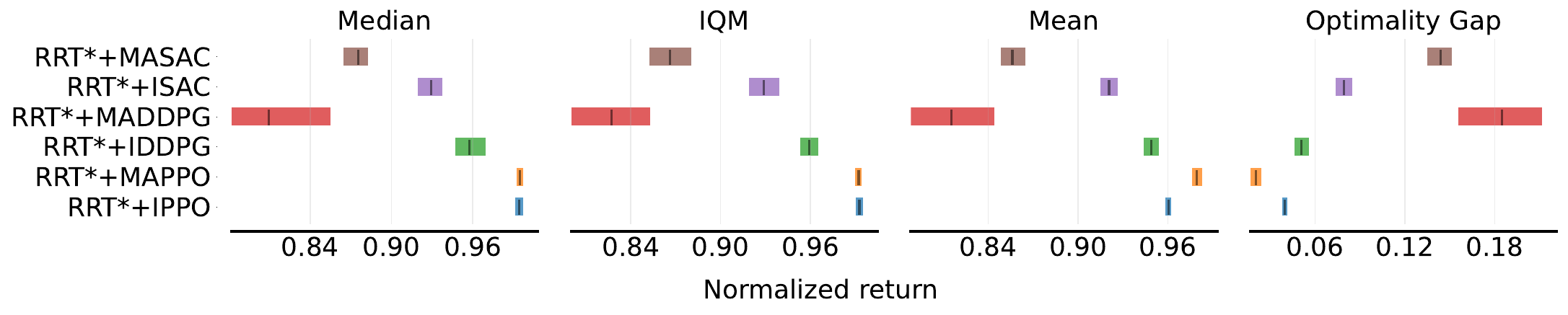}
    \end{subfigure}
    \caption{Normalized return scores. Top: MARL baselines. Bottom: RRT* hybrid methods. Metrics include median, interquartile mean (IQM), mean, and optimality gap. Higher values are better except for optimality gap. Each bar shows 95\% confidence intervals.}
    \label{fig:aggregate_scores}
\end{figure*}

We also report aggregate return metrics across all evaluated methods in Fig.\ref{fig:aggregate_scores}. These charts, produced using the \textsc{marl-eval} toolkit\cite{gorsane2022towards}, show median, interquartile mean (IQM), mean return, and optimality gap, with normalized scores. Among MARL baselines, MAPPO performs best across all metrics, followed by IPPO. ISAC and IDDPG achieve lower scores and are less consistent, while MADDPG performs the worst and shows the largest optimality gap. Among hybrid methods, RRT*+MAPPO and RRT*+IPPO achieve strong results that match or exceed their baselines. However, RRT*+MASAC and RRT*+MADDPG perform poorly. This may be caused by the larger input space, especially for centralized critics that struggle with high-dimensional observations.

To further evaluate consistency and reliability, we present performance profiles in Fig.\ref{fig:performance_profile_return} and Fig.\ref{fig:performance_profile_success_rate}. These plots show the fraction of evaluation runs where each algorithm achieves a score above a given threshold. The results confirm our earlier findings: MAPPO and IPPO maintain good performance across tasks, while MADDPG and MASAC drop quickly. RRT*+MAPPO and RRT*+IPPO remain strong among hybrid approaches, while RRT*+MADDPG and RRT*+MASAC again show poor results.

These charts together highlight the differences in generalization ability, robustness, and effectiveness across MARL and hybrid methods. They also support our earlier analysis of SR, FT, MS, CO (Table~\ref{tab:performance_comparison}). Overall, these findings show that RRT* integration can improve performance for some methods, but not all. The effect depends on the algorithm and how it handles the added input complexity.

\paragraph{Sample Efficiency.}
\begin{figure}[htb!]
    \centering
    \includegraphics[width=\linewidth]{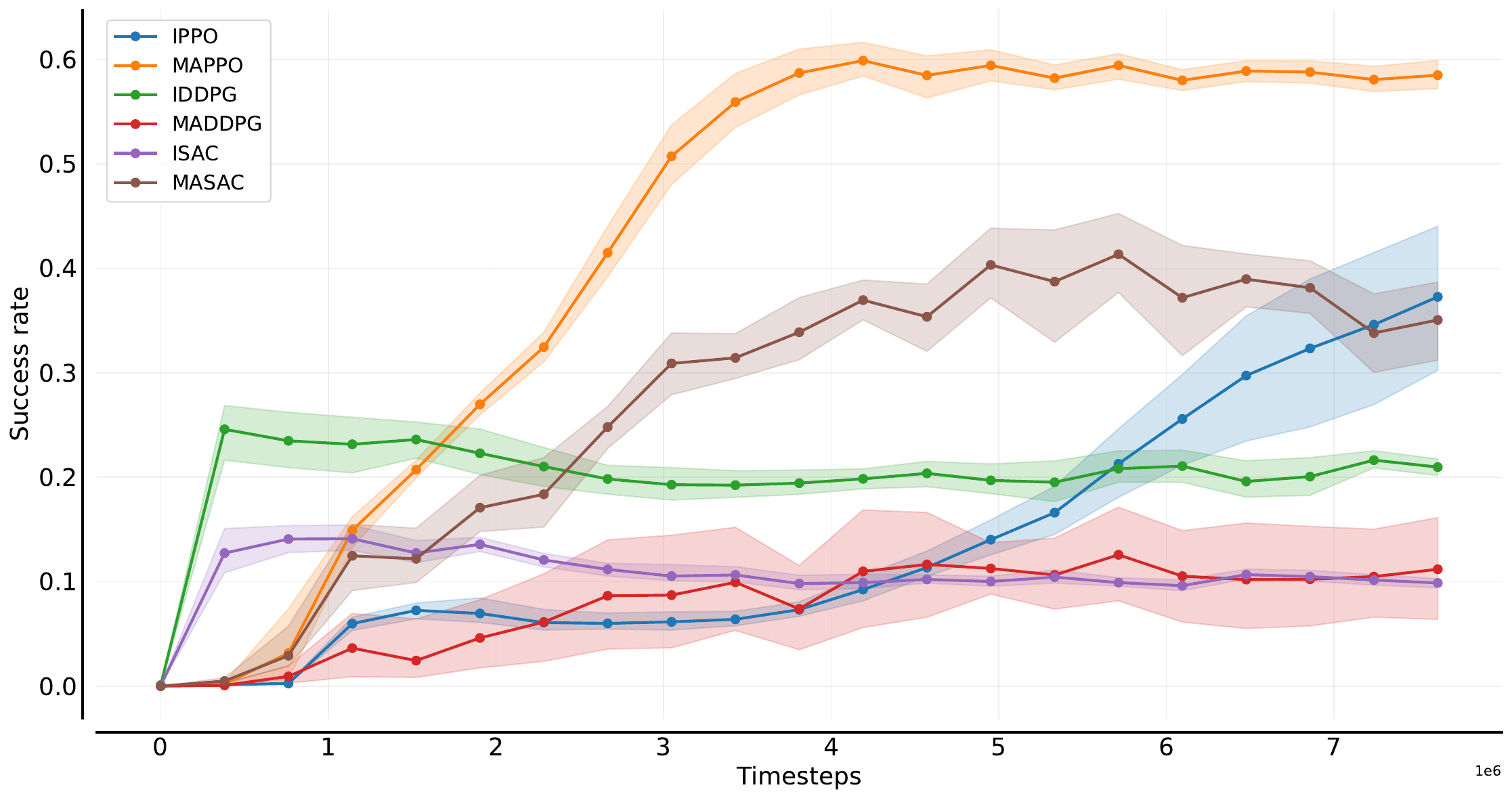}
    \caption{Sample-efficiency curves: mean success rate versus environment steps. Computed using \textsc{marl-eval}.}
    \label{fig:sample_efficiency}
\end{figure}

Figure~\ref{fig:sample_efficiency} shows the mean success rate evolving during training, computed with the \textsc{marl-eval} toolkit~\cite{gorsane2022towards}.
MAPPO learns the fastest, rising above 0.60 after roughly $3M$ steps and then remaining stable.  
MASAC improves quickly at first but plateaus below 0.45.  
IPPO starts slowly yet keeps improving and nearly reaches the MASAC curve near the end of training.  
IDDPG attains a modest plateau early and then changes little.  
ISAC climbs slightly in the first million steps and then flattens out, while MADDPG stays under 0.15 for the entire run.  

\paragraph{DDPG and SAC analysis}

\begin{table}[htb!]
\centering
\scriptsize
\begin{tabular}{lccccc}
\toprule
Algorithm & SR $\uparrow$ & FT $\downarrow$ & MS $\downarrow$ & CO $\uparrow$ & Ep. Return $\uparrow$ \\
\midrule
IDDPG (SC)   & 0.77 & 406  & 134 & 0.996 & 266  \\
ISAC (SC)    & 0.21 & 1060 & 160 & 0.999 & 92   \\
MADDPG (SC)  & 0.00 & 1241 & 160 & 0.997 & -1.1 \\
MADDPG (NSC) & 0.00 & 1259 & 160 & 0.998 & -0.9 \\
MASAC (SC)   & 0.26 & 813  & 159 & 0.996 & 67   \\
MASAC (NSC)  & 0.44 & 780  & 160 & 1.000 & 193  \\
\bottomrule
\end{tabular}
\caption{Performance comparison of multi-agent DDPG and SAC variants on \texttt{random\_grid} with 8 agents. SC: shared critic, NSC: not shared critic. }
\label{tab:ddpg_sac_comparison}
\end{table}

We further investigated the weak performance of multi-agent DDPG and SAC algorithms in our benchmark. Specifically, we observed that both MADDPG and MASAC achieved lower success rates and higher episode lengths compared to their independent counterparts (IDDPG and ISAC). To understand these results, we considered two possible factors that could negatively affect learning with centralized critics.

First, the credit assignment problem becomes harder as the number of agents increases. In both MADDPG and MASAC, a single centralized critic must estimate value based on the combined actions and states of all agents. The critic’s target is a simple mean across agents in this version, which can blur important agent-specific differences. This makes learning slow and unstable.

Second, the input size of the centralized critic grows with the number of agents. For example, in our setting with 8 agents, the critic receives a long concatenated vector of all observations and actions. This high-dimensional input makes the learning problem more difficult, especially for relatively small networks like MLPs. Previously discussed performance of RRT*+MASAC and RRT*+MADDPG also supports this hypothesis. These hybrid methods show much lower returns and success rates compared to RRT*+IPPO or RRT*+MAPPO. This drop in performance is likely caused by the added RRT* information increasing the input size even further, making the learning task more difficult for centralized critics.

To test these ideas, we trained new versions of MADDPG and MASAC where the critic network is no longer shared between agents. Instead, each agent has its own critic (without parameter sharing), while still using centralized training. The results are shown in Table~\ref{tab:ddpg_sac_comparison}. For MADDPG, there was no improvement: both the shared and non-shared versions failed to learn the task, with success rate remaining at 0.0. This further supports the hypothesis that MADDPG struggles with large input vectors and credit assignment, even when the critics are separated. In contrast, MASAC improved with non-shared critics: the success rate rose from 0.26 to 0.44 and return nearly tripled. This suggests that SAC can better tolerate large input sizes, likely due to its entropy regularization and smoother policy updates. However, even this improved MASAC variant still performs worse than ISAC, which does not face centralized credit assignment at all. This shows that credit assignment remains a challenge in MASAC, despite its robustness to large inputs. Notably, the poor performance of RRT*+MADDPG and RRT*+MASAC in earlier experiments aligns with these findings.

Together, these findings confirm that both factors—the size of the critic input and the difficulty of shared credit assignment—can reduce the performance of centralized off-policy MARL methods. We include these results in Table~\ref{tab:ddpg_sac_comparison}.

\subsection{Benchmark Maps}

CAMAR includes three benchmark map types designed to evaluate navigation and coordination in multi-agent scenarios. These maps differ in layout, complexity, and agent configurations.

\paragraph{\texttt{random\_grid}} map is a $20 \times 20$ grid with randomly placed rectangular obstacles (with circle-based discretization). Agents are initialized in free cells, and goals are placed at randomly chosen locations. The scenario uses holonomic dynamics and includes parameters for obstacle density, agent radius, goal radius, and others. During benchmarking, we generate six different \texttt{random\_grid} instances: three maps with 8 agents and three with 32 agents. Each of these variations is evaluated under three different obstacle densities (0.0, 0.05, 0.15), which define how cluttered the map becomes. These tasks are relatively simple but still useful to test multi-agent pathfinding (MAPF) algorithms.

\paragraph{\texttt{labmaze\_grid}} is based on procedural maze generation. It creates more structured maps using multiple rectangular rooms and corridors. The parameter \texttt{extra\_connection\_probability} controls how many connections exist between rooms: 1.0 means the map is fully connected, while lower values introduce more isolated areas and bottlenecks. As with the random grid, we evaluate 6 instances: three maps with 8 agents and three with 32 agents, each under three different connection probabilities (0.4, 0.65, 1.0). The dynamics are the same as in the random grid, allowing consistent comparison across layouts. These tasks provide harder challenges for coordination and navigation.

These benchmark maps are scalable, fast to simulate, and allow flexible control over structure and difficulty. Together, they form a comprehensive testbed for MAPF.

\subsection{Heterogeneous Support}

The \texttt{hetero\_give\_way} map tests heterogeneous multi-agent coordination. It is based on a simple corridor scenario where one larger agent cannot enter the central chamber, while the smaller agent can pass through. This setup forces agents to learn implicit turn-taking or yielding behavior. Although the dynamics remain the same (\texttt{HolonomicDynamic}), this configuration highlights differences in agent capabilities. Basic algorithms like MAPPO and IPPO struggle in this task. Nonetheless, CAMAR supports such heterogeneous settings out of the box and can be used to advance research in coordination strategies for agents with diverse sizes, dynamics, and behaviors.

\section{Appendix E --- Extended Scalability Analysis}
\label{app:scalability}

\paragraph{CAMAR vs VMAS}

\begin{figure}[htb!]
    \centering
    \begin{minipage}[b]{90pt}
        \centering
        \includegraphics[clip,trim={0pt 0pt 0pt 0pt}, width=80pt]{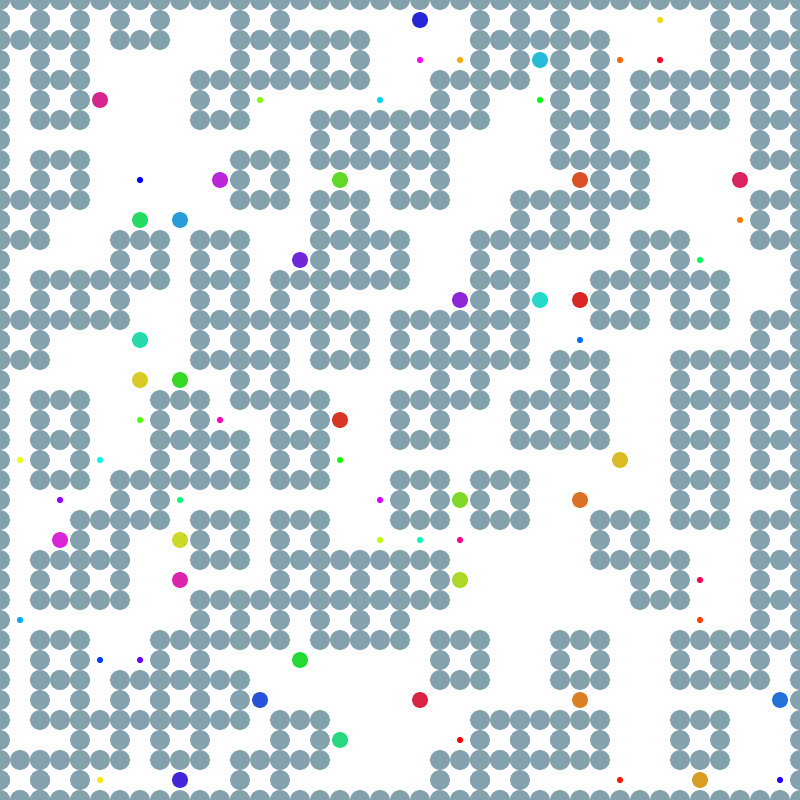}
        \subcaption{CAMAR}
        \label{subfig:camar_random_grid}
        \vspace{8pt}

        \includegraphics[clip,trim={13pt 13pt 13pt 13pt}, width=80pt]{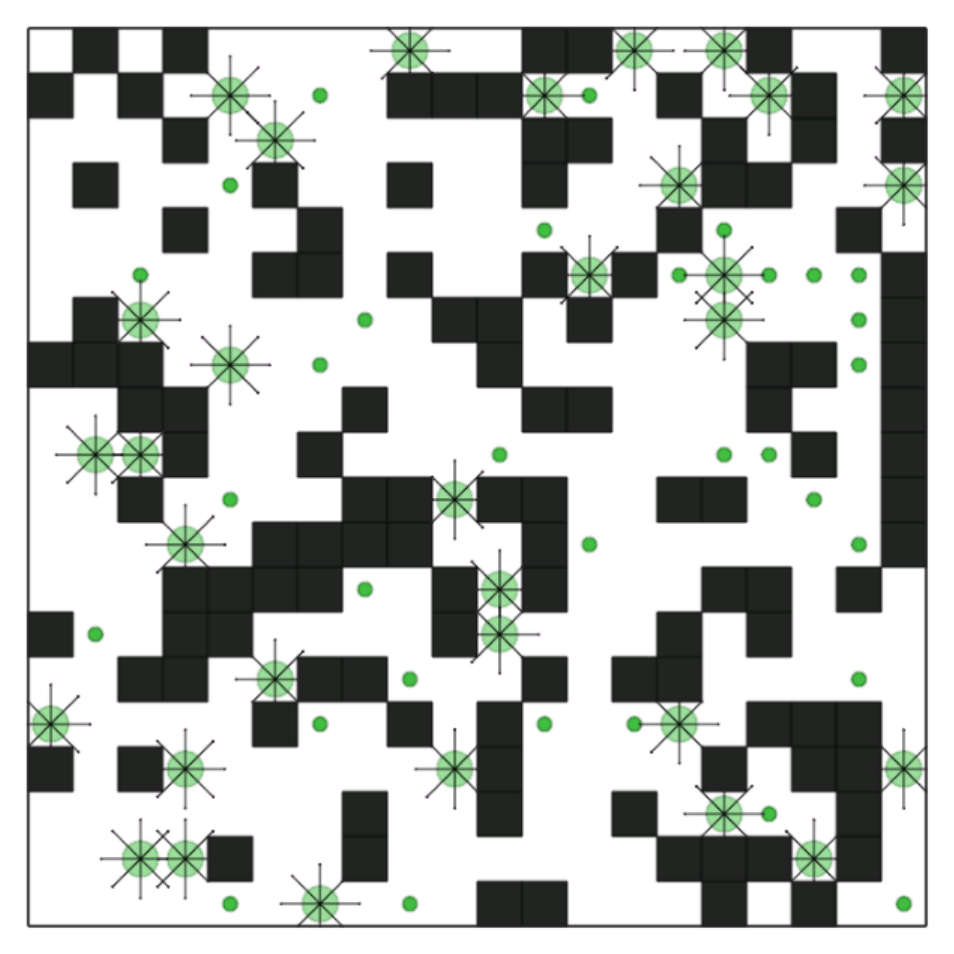}
        \subcaption{VMAS}
        \label{subfig:vmas_random_grid}
        
        \vspace{8pt}
    \end{minipage}
    \begin{minipage}[b]{130pt}
        \centering
        \includegraphics[clip,trim={0pt 0pt 0pt 0pt}, width=130pt]{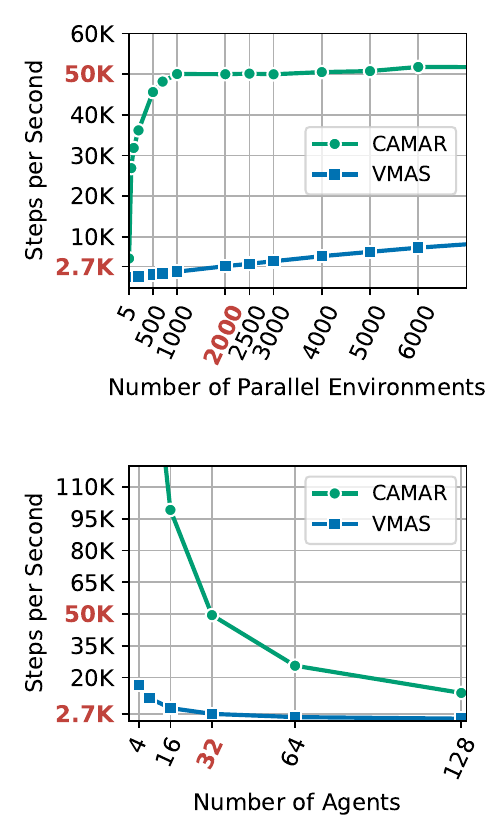}
    \end{minipage}
    \vspace{5px}
    \caption{Scalability comparison between CAMAR and VMAS across different settings, evaluating the impact of increasing parallel environments, agent count, and obstacles on simulation speed.}
    \label{fig:scalability_analysis2}
\end{figure}

This section provides visualizations of scenarios of CAMAR and VMAS. Fig.~\ref{subfig:scalability_analysis_a} and Fig.~\ref{fig:scalability_analysis2} use the same map generation settings for both CAMAR and VMAS to allow a fair comparison under matched conditions.

\paragraph{Comparing with Other Environments.} In Table~\ref{tab:scalabilty_comparison}, we compare CAMAR to other environments with GPU support like JaxMARL and Jumanji. CAMAR is faster than VMAS and the MPE baselines, and it reaches speeds close to the best GPU simulators. For example, CAMAR runs at 338K SPS with 4 agents and 50K SPS with 32 agents. Some JaxMARL environments reach higher throughput—for example, SMAX achieves up to 1.2M SPS—but their performance drops more quickly as the number of agents grows. JaxNav, for instance, cannot run with more than 32 agents. While SMAX and RWARE show strong throughput, they use simplified settings like open space or discrete grids and do not support procedurally generated obstacle layouts. CAMAR, in contrast, supports continuous control and dense obstacles, which brings more realism. It trades a small drop in speed for a richer environment that better matches real-world MAPF challenges.

\begin{table}[bth!]
\centering
\scriptsize
\begin{tabular}{lcccccc}
\toprule
Agents    & 4      & 8      & 16     & 32     & 64       & 128      \\
\midrule
CAMAR     & 338K   & 189K   & 99K    & 50K    & 25K      & 13K      \\
VMAS      & 16K    & 10K    & 5.5K   & 2.7K   & 1.2K     & 447      \\
JaxNav    & 159K   & 61K    & 13.6K  & 353    & -        & -        \\
SMAX      & 0.8M   & 1.0M   & 1.2M   & 0.8M   & 0.3M     & 0.1M     \\
RWARE     & 1.4M   & 1.1M   & 0.8M   & 0.5M   & 0.3M     & 0.1M     \\
MPE-2     & 0.6M   & 0.8M   & 1.2M   & 0.7M   & 0.5M     & 0.2M     \\
MPE-1120  & 11.8K  & 11.2K  & 11.9K  & 11.1K  & 9.7K     & 7.7K     \\
\bottomrule
\end{tabular}
\caption{SPS vs number of agents. Comparison across environments with GPU support. MPE-2 uses the default simple-tag configuration with 2 circle obstacles, while MPE-1120 uses 1120 circles matching the number of circle obstacles in CAMAR. `-` indicates that the environment was too slow to even finish episode.}
\label{tab:scalabilty_comparison}
\end{table}

\section{Appendix F --- Future Work}
\label{app:future}

We highlight several research directions that can be explored using CAMAR:

\begin{enumerate}
    \item CAMAR enables testing with large agent populations, supporting the development of scalable multi-agent reinforcement learning (MARL) methods. Future work can focus on improving how algorithms handle hundreds of agents operating in dense, dynamic environments.
    \item CAMAR provides a natural platform for studying inter-agent communication. Large-scale coordination often requires agents to exchange information with nearby teammates, which remains an open challenge. The environment offers the necessary tools to explore such localized communication mechanisms.
    \item CAMAR includes realistic continuous navigation tasks, making it suitable for integrating classical planning approaches (RRT, RRT$^*$) with learning-based agents. Combining long-term planning with reinforcement learning could improve both sample efficiency and robustness.
    \item CAMAR supports agents with different dynamics and properties, enabling research on heterogeneous teams. We plan to extend the current list of dynamics. For the moment, we include differential-drive and holonomic robots, but we are going to support more options, such as car-like and 2D quadrotor dynamics.
    \item CAMAR’s fully modular design allows users to easily redefine observation and reward functions through wrapping. We plan to introduce additional observation types, including simulated 2D LiDAR, to further diversify sensing modalities for navigation tasks.
    \item Another direction of future extensions is to extend classical baselines beyond the current RRT+PD and RRT$^*$ setup by adding widely used multi-robot navigation methods such as PRM+PD (Probabilistic roadmap for global planning with PD controller), ORCA, and MPPI. Similarly, we will expand learning-based baselines to include RNN-based algorithms and architectures using graph attention or transformers, going beyond the standard MLP-based backbones.
    \item We aim to incorporate post-generation connectivity checks to ensure that all agents can reach their goals. While computationally expensive, this process will improve the reliability and fairness of evaluation.
    \item CAMAR can be extended beyond multi-agent pathfinding. Future releases will include new task types such as pick-up-and-delivery, cooperative transportation, and dynamic obstacle scenarios, making CAMAR more general testbed for large-scale MARL research.
\end{enumerate}

We hope CAMAR will support the community in studying these and other open problems in multi-agent learning and planning.

\section{Appendix G --- Implementation Details}
\label{app:implementation}

\subsection{Integrations}
CAMAR is designed to work easily with modern reinforcement learning tools. It follows the Gymnax interface, which is already familiar to many researchers working with RL environments on top of JAX~\cite{jax2018github}.

We also provide a wrapper for TorchRL~\cite{bou2023torchrl}, which allows users to integrate CAMAR into PyTorch-based pipelines. In addition, CAMAR supports integration with BenchMARL~\cite{bettini2024benchmarl}, a framework for evaluating MARL algorithms. These integrations make it easy to use CAMAR with popular RL frameworks.

\subsection{Vectorized setup}
To maintain high simulation speed, CAMAR uses efficient vectorized operations based on JAX ~\cite{jax2018github}. Agents with the same dynamic model are grouped together and updated in parallel. For agent sizes, two cases are supported: if all agents have the same radius, it is treated as a constant during JAX compilation ~\cite{jax2018github}, giving faster simulation. If agents have different sizes, their radii are passed as vectors being part of the environment state, which still allows efficient processing and supports randomized sizes during training.

\section{Appendix H --- Extended Related Work}
\label{app:related}

Many multi-agent reinforcement learning (MARL) benchmarks have been developed in recent years, each focusing on different challenges such as coordination, navigation, or multi-agent planning. A key ability for agents in robotics is to move and adapt in complex environments. Some recent environments explore these abilities, but they differ widely in important properties.

These differences include whether the environment supports continuous spaces, GPU acceleration, and high simulation throughput. Some platforms support only discrete actions, while others offer more realistic continuous control. Environments also vary in their ability to handle large groups of agents or support partial observability.

Other practical features also matter for researchers. Some environments are fully implemented in Python and support flexible task generation, which makes them easier to extend and use in large experiments. Others may lack documentation or require heavy customization.

Finally, only a few MARL environments offer standard evaluation protocols, automated tests, or package installation via PyPi. These features are useful for ensuring fair comparisons between methods, improving reproducibility, and making environments easier to maintain. In our work, we aim to address these points by extending existing protocols and providing a high-performance, flexible platform.

To sum up, our benchmark builds on the strengths of existing environments and aims to complement ongoing efforts in MARL research. Over the years, a range of benchmarks has helped researchers tackle challenges in coordination, planning, and navigation. Well-known examples include SMAC~\cite{samvelyan2019starcraft, ellis2023smacv2, rutherford2023jaxmarl}, Jumanji~\cite{bonnet2023jumanji}, POGEMA~\cite{skrynnik2022pogema, skrynnik2025pogema}, MPE~\cite{mordatch2017emergence, lowe2017multi}, and VMAS, each of which introduces useful features for different kinds of tasks.

SMAC~\cite{samvelyan2019starcraft, ellis2023smacv2, rutherford2023jaxmarl} is popular for testing strategic decision-making, but it uses discrete actions and does not scale well for large environments. Jumanji~\cite{bonnet2023jumanji} supports GPU acceleration and procedural generation, but its focus is not on navigation or planning. POGEMA~\cite{skrynnik2022pogema, skrynnik2025pogema} is strong in grid-based navigation and procedural tasks with many agents, but it does not use continuous states or actions, which are important in robotics.

MPE~\cite{mordatch2017emergence, lowe2017multi} has played an important role in early MARL research, but it cannot scale to hundreds of agents efficiently. VMAS~\cite{bettini2022vmas} builds on MPE~\cite{mordatch2017emergence, lowe2017multi} by adding its physics and continuous dynamics, making it better for robotics. However, it can still be slow and difficult to scale to larger agent populations or complex maps.

Many environments also lack evaluation protocols. This makes it difficult to compare algorithms fairly and limits the reproducibility of results. Performance issues also appear in environments that depend heavily on CPU-GPU communication, which slows down training and makes large-scale experiments harder.

\begin{table*}[tb!]
\centering 
\small
\rowcolors{2}{gray!15}{white}
\begin{tabular}{p{3.8cm}p{0.6cm}p{0.3cm}p{0.3cm}p{0.3cm}p{0.3cm}p{0.3cm}p{0.3cm}p{0.88cm}p{0.3cm}p{0.3cm}p{0.88cm}p{0.3cm}p{0.3cm}p{0.3cm}}
Environment / Simulator & \rotlabel{Repository} & \rotlabel{Cont. Observations} & \rotlabel{Cont. Actions} & \rotlabel{GPU Support} & \rotlabel{Scalability \textgreater{}500 Agents} & \rotlabel{Partially observable} & \rotlabel{Heterogeneous agents} & \rotlabel{Performance \textgreater{}10K SPS}  & \rotlabel{Python based} & \rotlabel{Procedural generation} & \rotlabel{Requires generalization} & \rotlabel{Evaluation protocols} & \rotlabel{Tests \& CI} & \rotlabel{PyPI Listed} \\
\midrule

Waterworld (SISL)~\cite{gupta2017cooperative} & \href{https://pettingzoo.farama.org/environments/sisl/}{link} & \cm & \cm & \xm & \xm & \cm & \xm & \cm & \cm & \xm & \xm & \xm & \cm & \cm \\

RWare~\cite{papoudakis2020benchmarking} & \href{https://github.com/sisl/MADRL.git}{link} & \xm & \xm & \xm & \xm & \cm & \xm & \xm & \cm & \xm & \cm & \xm & \cm & \cm \\

RWare (Jumanji)~\cite{bonnet2023jumanji} & \href{https://github.com/instadeepai/jumanji.git}{link} & \xm & \xm & \cm & \xm & \cm & \xm & \xm & \cm & \xm & \cm & \xm & \cm & \cm \\

RWare (Pufferlib)~\cite{suarez2024pufferlib} & \href{https://github.com/PufferAI/PufferLib.git}{link} & \xm & \xm & \xm & \cm & \cm & \xm & \cm & \xm & \xm & \cm & \xm & \cm & \cm \\

Trash Pickup (Pufferlib)~\cite{suarez2024pufferlib} & \href{https://github.com/PufferAI/PufferLib.git}{link} & \xm & \xm & \xm & \cm & \cm & \xm & \cm & \xm & \xm & \cm & \xm & \cm & \cm \\

SMAC~\cite{samvelyan2019starcraft} & \href{https://github.com/oxwhirl/smac}{link} & \cm & \xm & \xm & \xm & \cm & \cm & \xm & \xm & \xm & \xm & \xm & \xm & \xm \\

SMACv2~\cite{ellis2023smacv2} & \href{https://github.com/oxwhirl/smacv2.git}{link} & \cm & \xm & \xm & \xm & \cm & \cm & \xm & \xm & \xm & \cm & \xm & \xm & \xm \\

SMAX (JaxMARL)~\cite{rutherford2023jaxmarl} & \href{https://github.com/FLAIROx/JaxMARL.git}{link} & \cm & \cm & \cm & \xm & \cm & \cm & \cm & \cm & \xm & \cm & \xm & \cm & \cm \\

MPE ~\cite{mordatch2017emergence, lowe2017multi} & \href{https://github.com/Farama-Foundation/MPE2.git}{link} & \cm & \cm & \xm & \cm & \cm & \cm & \xm\ /\ \cm$^{~\ref{ff:sps}}$ & \cm & \xm & \cm & \xm & \cm & \cm \\

MPE (JaxMARL)~\cite{rutherford2023jaxmarl} & \href{https://github.com/FLAIROx/JaxMARL.git}{link} & \cm & \cm & \cm & \cm & \cm & \cm & \xm\ /\ \cm$^{~\ref{ff:sps}}$ & \cm & \xm & \cm & \xm & \cm & \cm \\

JaxNav (JaxMARL)~\cite{rutherford2024no} & \href{https://github.com/FLAIROx/JaxMARL.git}{link} & \cm & \cm & \cm & \xm & \cm & \xm & \xm & \cm & \xm & \xm & \xm & \cm & \cm \\

Nocturne~\cite{vinitsky2022nocturne} & \href{https://github.com/facebookresearch/nocturne.git}{link} & \cm & \cm & \xm & \cm & \cm & \xm & \xm & \xm & \xm & \xm & \cm & \cm & \xm \\

POGEMA~\cite{skrynnik2022pogema} & \href{https://github.com/CognitiveAISystems/pogema.git}{link} & \xm & \xm & \xm & \cm & \cm & \xm & \cm & \cm & \cm & \cm & \cm & \cm & \cm \\

VMAS$^{~\ref{ff:vmas}}$~\cite{bettini2022vmas} & \href{https://github.com/proroklab/VectorizedMultiAgentSimulator.git}{link} & \cm & \cm & \cm & \xm & \cm & \cm & \xm\ /\ \cm$^{~\ref{ff:vmas}}$ & \cm & \xm & \xm\ /\ \cm$^{~\ref{ff:vmas}}$ & \xm & \cm & \cm \\

SMART ~\cite{yan2025advancing} & \href{https://github.com/JingtianYan/SMART.git}{link} & \cm & \cm & \xm & \cm & \cm & \xm & \xm & \xm & \xm & \cm & \xm & \xm & \xm \\

\midrule

Gazebo ~\cite{koenig2004design} & \href{https://gazebosim.org/home}{link} & \cm & \cm & \cm & \xm & \cm & \cm & \xm & \xm & \xm & \xm & \xm & \cm & \xm \\

Webots ~\cite{michel2004cyberbotics} & \href{https://cyberbotics.com/}{link} & \cm & \cm & \cm & \xm & \cm & \cm & \xm & \xm & \xm & \xm & \xm & \cm & \xm \\

ARGoS ~\cite{pinciroli2012argos} & \href{https://www.argos-sim.info/}{link} & \cm & \cm & \xm & \cm & \cm & \cm & \xm & \xm & \xm & \xm & \xm & \cm & \xm \\

\midrule

CAMAR (Ours) & \href{https://github.com/AIRI-Institute/CAMAR.git}{link} & \cm & \cm & \cm & \cm & \cm & \cm & \cm & \cm & \cm & \cm & \cm & \cm & \cm \\

\bottomrule
\end{tabular}
\caption{Comparison of multi-agent reinforcement learning (MARL) environments and simulators. Each row corresponds to a specific environment or a particular implementation of it. The columns indicate key properties, including support for continuous observations and actions, GPU acceleration, scalability beyond 500 agents, partial observability, heterogeneous agents, and whether the simulator can exceed 10K simulation steps per second (SPS). Additional columns specify if the environment is implemented fully in Python, supports procedural generation, requires generalization across different maps or tasks, includes evaluation protocols, and provides built-in tests or continuous integration (CI). The “Repository” column contains links to the official source code for each environment. CAMAR, listed at the bottom, is our proposed environment.}
\label{tab:marl-environments-det}
\end{table*}

Finally, popular robotics simulators such as Gazebo~\cite{koenig2004design}, Webots~\cite{michel2004cyberbotics}, and ARGoS~\cite{pinciroli2012argos} are powerful tools for realistic physics-based simulation and robotic deployment. Each of these frameworks serves specific use cases focused on accurate physical modeling and hardware-oriented evaluation. In contrast, our environment, CAMAR, is designed to complement these tools by focusing on large-scale MARL and benchmarking, particularly for continuous MAPF tasks. Rather than replacing existing simulators, CAMAR extends the set of available research tools, bridging the gap between large-scale MARL studies and robotics-oriented simulation.

We compare MARL environments and robotic simulators using the following criteria (Tables~\ref{tab:marl-environments},~\ref{tab:marl-environments-det}):

\paragraph{Continuous Observations and Actions}
Robots usually operate in continuous state and action spaces, so it's important for benchmarks to reflect these conditions. Environments like VMAS and MPE~\cite{mordatch2017emergence, lowe2017multi} use continuous actions and observations, while others, like Trash Pickup~\cite{suarez2024pufferlib} or POGEMA~\cite{skrynnik2025pogema}, use discrete representations, which limit realism in robotics simulations.

\paragraph{GPU Support}

In multi-agent environments, each agent often has its observations and rewards. This can lead to heavy data transfers between the CPU and the GPU, especially when training deep RL models. Benchmarks like Jumanji~\cite{bonnet2023jumanji} and SMAX~\cite{rutherford2023jaxmarl} are initially built with GPU acceleration, helping to reduce training time. In contrast, many older environments like SMAC~\cite{samvelyan2019starcraft, ellis2023smacv2} and MPE~\cite{mordatch2017emergence, lowe2017multi} do not support GPU-based simulation.

\paragraph{Scalability \textgreater{}500 Agents}
When the number of agents grows, decision-making becomes harder. A good benchmark should scale well to hundreds or thousands of agents. Environments like POGEMA~\cite{skrynnik2022pogema, skrynnik2025pogema} and Trash Pickup~\cite{suarez2024pufferlib} can run with thousands and even millions of agents. Others, like VMAS, MPE~\cite{mordatch2017emergence, lowe2017multi}, and SMAC~\cite{samvelyan2019starcraft, ellis2023smacv2}, are limited to much smaller groups.

\paragraph{Partially Observable}
In most real-world scenarios, agents see only part of the environment. This is known as partial observability and is a common feature in almost every RL environment. SMAC~\cite{samvelyan2019starcraft, ellis2023smacv2}, RWare~\cite{papoudakis2020benchmarking}, and POGEMA~\cite{skrynnik2022pogema, skrynnik2025pogema} support this feature by limiting each agent’s view.

\paragraph{Heterogeneous Agents}
In real life, agents and robots can have different sensors, shapes, or goals. Benchmarks like VMAS and SMAC~\cite{samvelyan2019starcraft, ellis2023smacv2} allow agents to behave differently, making cooperation more complex. Others, like POGEMA~\cite{skrynnik2022pogema, skrynnik2025pogema} or RWare~\cite{papoudakis2020benchmarking}, usually involve homogeneous agents.

\paragraph{Performance \textgreater{}10K Steps/s}
High simulation speed is essential when agents need millions of interactions to learn. For example, the Pufferlib~\cite{suarez2024pufferlib} environment can run with a speed of up to 1M steps per second. This allows researchers to train and develop models more quickly. Environments like SMAC~\cite{samvelyan2019starcraft, ellis2023smacv2} or VMAS can be slower, especially when running with many agents.

\paragraph{Python Based}
Using Python makes it easier for researchers to understand and modify the environment. SMAX~\cite{rutherford2023jaxmarl}, VMAS, and POGEMA~\cite{skrynnik2022pogema, skrynnik2025pogema} are implemented entirely in Python. This helps with faster development and integration into learning frameworks.

\paragraph{Procedural Generation}
Procedural generation helps create diverse tasks that reduce the chance of overfitting. POGEMA~\cite{skrynnik2022pogema, skrynnik2025pogema} and Jumanji~\cite{bonnet2023jumanji} use this method to generate complex tasks automatically. Other environments, like RWare~\cite{papoudakis2020benchmarking}, use fixed layouts that limit diversity.

\paragraph{Requires Generalization}
Only a few environments offer separate training and testing tasks, making it difficult to test whether agents generalize well to new situations. POGEMA~\cite{skrynnik2022pogema, skrynnik2025pogema} and SMACv2~\cite{ellis2023smacv2} include test scenarios that allow researchers to check generalization. Many others use the same tasks during both training and testing.

\paragraph{Evaluation Protocols}
To compare algorithms fairly, we need well-defined test cases and metrics. Benchmarks like SMACv2~\cite{ellis2023smacv2} and Jumanji~\cite{bonnet2023jumanji} include evaluation protocols to make results reproducible and meaningful. Many other environments do not have standard evaluation tools or test setups.

\paragraph{Tests \& CI}
CI pipeline and testing suite are vital for collaborative development and maintenance of open-source projects. These practices help ensure code reliability and facilitate contributions from the whole research community.

\paragraph{PyPi Listed}
When an environment is available on PyPi, it becomes easier for others to install and use. Benchmarks like Jumanji~\cite{bonnet2023jumanji}, and Trash Pickup~\cite{suarez2024pufferlib} are available on PyPi. This lowers the barrier to entry and helps increase adoption in the research community.

Here are detailed descriptions and analysis of each environment used for the comparison in the Tables~\ref{tab:marl-environments},~\ref{tab:marl-environments-det}:

\paragraph{Waterworld (SISL)} — The Waterworld environment, part of the SISL (Stanford Intelligent Systems Laboratory) suite ~\cite{gupta2017cooperative}, is a continuous control benchmark where multiple agents move in a bounded two-dimensional space to collect moving targets (“food”) while avoiding harmful objects (“poison”). Agents have continuous observations and actions, and their motion dynamics are simple and lightweight, which allows high simulation speeds for training.

The environment places a small number of obstacles randomly in the scene, with the default setup containing only one obstacle. This limited variation does not require agents to generalize across different maps in a meaningful way. Waterworld does not support heterogeneous agents, evaluation protocols, or complex multi-stage tasks, and the scenario remains structurally the same across runs.

\paragraph{Multi-Robot Warehouse (RWare)} — The Multi-Robot Warehouse environment~\cite{papoudakis2020benchmarking} simulates robots moving in a warehouse to collect and deliver requested goods. The layout is grid-based, and agents must coordinate to navigate around shelves and other robots. The default version is implemented in Python using a discrete state and action space. It supports partial observability but does not provide procedural generation: the warehouse layout is fixed. As a result, agents do not need to generalize across different maps.

Several re-implementations of RWare exist. The Jumanji version~\cite{bonnet2023jumanji} rewrites the environment in JAX, which improves simulation speed and allows hardware acceleration on GPUs, but keeps the original fixed-layout design and task structure. The PufferLib version~\cite{suarez2024pufferlib} modifies the environment to support large-scale parallel simulation.

None of the RWare versions support heterogeneous agents, continuous control, or evaluation protocols. Despite this, RWare remains a widely used benchmark for discrete-space multi-agent pathfinding and cooperative delivery tasks.

\paragraph{Trash Pickup (PufferLib)} — The Trash Pickup environment~\cite{suarez2024pufferlib} is a grid-based multi-agent task where agents move around a map to collect pieces of trash and deliver them to designated drop-off locations. It is implemented in C with a Python API for controlling agents, which allows efficient simulation. The environment uses discrete state and action spaces and supports partial observability.

The layout is fixed, and trash positions follow a predefined spawn pattern. While trash locations may vary between episodes, the underlying map structure remains the same. This means the environment does not require generalization across different maps. The design supports scalability to more than 500 agents.

The Trash Pickup benchmark does not provide continuous control, heterogeneous agents, or formal evaluation protocols. However, it offers a simple and repeatable cooperative task that can be scaled to large numbers of agents, making it a useful test case for studying coordination efficiency in discrete environments.

\paragraph{StarCraft Multi-Agent Challenge (SMAC)} — The StarCraft Multi-Agent Challenge~\cite{samvelyan2019starcraft} is one of the most widely used benchmarks in the MARL community. In SMAC, a team of StarCraft II units controlled by independent agents must cooperate to defeat an opposing team. The environment is partially observable, and by default uses discrete action spaces. The underlying maps are fixed, which allows agents to solve the tasks without relying on observation inputs by simply memorizing the optimal action sequence for each map. This significantly reduces the need for generalization across scenarios.

\paragraph{SMACv2} — SMACv2~\cite{ellis2023smacv2} addresses one of the key limitations of SMAC by introducing randomly generated maps with random positions of units, which prevents memorization of fixed action sequences and forces agents to generalize their policies across start positions. The rest of the environment remains the same as SMAC, with partially observable states and discrete actions by default. Like SMAC, SMACv2 does not include an evaluation protocol, but it can be evaluated using the protocol proposed in~\cite{gorsane2022towards}.

\paragraph{SMAX (JaxMARL)} — SMAX~\cite{rutherford2023jaxmarl} is a JAX-based re-implementation of SMACv2 that leverages hardware acceleration for faster simulation. In addition to performance improvements, SMAX introduces the option to switch from discrete to continuous action spaces, making it more flexible for testing different types of MARL algorithms.

\paragraph{Multi-Agent Particle Environment (MPE)} — The Multi-Agent Particle Environment~\cite{mordatch2017emergence, lowe2017multi} is a lightweight 2D simulator for testing cooperative, competitive, and communication multi-agent tasks. Agents and landmarks are represented as simple geometric shapes, and their interactions follow basic physical dynamics. MPE supports both continuous and discrete action spaces, and scenarios can be either fully or partially observable. Procedural generation is not used in the default scenarios, although obstacle or landmark positions can be randomized in some tasks. This means generalization across different maps is not strictly required. The environment does not include a built-in evaluation protocol.

\paragraph{MPE (JaxMARL)} — The JaxMARL re-implementation of MPE~\cite{rutherford2023jaxmarl} offers hardware acceleration through JAX, which enables much faster simulation compared to the original Python implementation. Functionally, it mirrors MPE in terms of available scenarios, physics, and agent capabilities. It supports both continuous and discrete action spaces. Like the original MPE, it does not include procedural map generation by default and does not provide an evaluation protocol.

\paragraph{JaxNav (JaxMARL)} — JaxNav~\cite{rutherford2024no} is a navigation-focused benchmark implemented within the JaxMARL framework. It places agents in a continuous 2D space where they must navigate to goal locations while avoiding static obstacles. The environment supports both continuous and discrete action spaces. Maps are fixed in the default setup, so agents do not need to generalize to unseen layouts. JaxNav is implemented in JAX for hardware acceleration but does not provide an evaluation protocol.

\paragraph{Nocturne} — Nocturne \cite{vinitsky2022nocturne} is a 2D partially observed driving simulator implemented in C++ for performance, with a Python API for training and evaluation. It focuses on realistic autonomous driving scenarios sourced from the Waymo Open Dataset. The environment provides a set of benchmark tasks in which agents control autonomous vehicles interacting with traffic participants whose trajectories are taken from recorded data. This setup captures complex multi-agent interactions requiring coordination, prediction of other agents’ intentions, and handling of partial observability.

Nocturne does not use procedurally generated maps, as all scenes are drawn from fixed datasets. While the simulator benefits from its C++ implementation, it does not provide GPU acceleration, which limits its throughput for large-scale MARL experiments.

\paragraph{POGEMA} — The Partially Observable Grid Environment for Multiple Agents~\cite{skrynnik2022pogema} is a grid-based benchmark for partially observable multi-agent pathfinding. Agents operate with ego-centric local observations and must reach individual goals while avoiding collisions. A key feature of POGEMA is its procedural map generation, producing diverse layouts—random maps, mazes, and warehouse-style scenarios—that require agents to generalize to unseen environments. It also offers integrated evaluation protocols with standardized metrics, enabling consistent comparison across reinforcement learning, planning, and hybrid methods. Implemented in Python, POGEMA supports scalable multi-agent experiments with both classical and learning-based policies.

\paragraph{VMAS} — The Vectorized Multi-Agent Simulator for Collective Robot Learning~\cite{bettini2022vmas} is a PyTorch-based, vectorized 2D physics framework designed for efficient multi-robot benchmarking. It includes a modular interface for defining custom scenarios, alongside a set of built-in multi-robot tasks, that each involve a relatively small number of agents. VMAS stands out for its GPU-accelerated, batch-simulated environments. It supports inter-agent communication and sensors (e.g., LIDAR).

\paragraph{SMART} — Scalable Multi-Agent Realistic Testbed~\cite{yan2025advancing} is a physics-based simulator tailored for bridging Multi-Agent Path Finding (MAPF) algorithms and real-world performance. Designed for scalability, SMART supports simulation of thousands of agents, enabling evaluation of large-scale deployments. However, its simulation speed is relatively low (around 335 SPS for 100 agents), which can limit large-scale reinforcement learning experiments. The platform targets both academic researchers and industrial users who lack access to extensive physical robot fleets, offering a realistic environment to test MAPF performance under near-real conditions.

\paragraph{Gazebo} — Gazebo is a widely used open-source 3D robotics simulator that integrates tightly with ROS. It supports detailed physics simulation, realistic sensors (e.g., LIDAR, cameras), and high-quality rendering. While Gazebo allows multi-robot experimentation and moderate scalability, it does not support GPU acceleration for simulation logic - GPU use is limited to rendering and sensor visuals only. This limitation can make simulations of large multi-agent systems slow or resource-heavy.

\paragraph{Webots} — Webots is a commercial-grade 3D robotics simulator offering a wide array of built-in robot models, sensors, and actuator libraries. It provides realistic physics simulation, high-quality graphics, and ROS support. Webots is well-suited for prototyping and academic or industrial robotic research. However, from a multi-agent perspective, it's not optimized for running large numbers of agents at scale—simulation speed tends to drop significantly as agent count grows, making massive multi-robot evaluations challenging.

\paragraph{ARGoS} — ARGoS~\cite{pinciroli2012argos} is an open-source, modular simulator designed specifically for swarm robotics and large-scale multi-agent systems. Its architecture supports running thousands of simple agents efficiently by combining multiple physics engines and leveraging parallel computation. ARGoS is customizable, allowing researchers to define their own robots, sensors, actuators, etc.

\end{document}